\documentclass{article} 
\usepackage{iclr2026_conference,times}

\usepackage{amsmath,amsfonts,bm}

\def\eqref#1{equation~\ref{#1}}

\def\1{\bm{1}}

\DeclareMathAlphabet{\mathsfit}{\encodingdefault}{\sfdefault}{m}{sl}
\SetMathAlphabet{\mathsfit}{bold}{\encodingdefault}{\sfdefault}{bx}{n}

\usepackage{hyperref}
\usepackage{url}
\usepackage[utf8]{inputenc} 
\usepackage[T1]{fontenc}    
\usepackage{booktabs}       
\usepackage{amsfonts}       
\usepackage{nicefrac}       
\usepackage{microtype}      
\usepackage{amsmath,amssymb} 
\usepackage{paralist}
\usepackage{comment}
\usepackage{adjustbox}
\usepackage{xkeyval}
\usepackage{calc}
\usepackage{colortbl}
\usepackage[dvipsnames]{xcolor}        
\usepackage{wrapfig}
\usepackage{nicefrac}       
\usepackage{tabularx}
\usepackage[normalem]{ulem}
\usepackage{bm}
\usepackage{bbm}
\usepackage{graphicx}
\usepackage{arydshln}
\usepackage{xspace}
\usepackage{enumitem}
\usepackage{multirow}
\usepackage{algorithm}
\usepackage{algorithmic}
\usepackage{soul}
\usepackage{pifont}

\def\eg{\emph{e}.\emph{g}.}
\def\ie{\emph{i}.\emph{e}.}

\usepackage[capitalize]{cleveref}
\crefname{section}{Sec.}{Secs.}
\Crefname{section}{Section}{Sections}
\Crefname{table}{Table}{Tables}
\crefname{table}{Tab.}{Tabs.}

\newcommand{\cmark}{\ding{51}}%
\newcommand{\xmark}{\ding{55}}%

\definecolor{hlgreen}{HTML}{C0F4D6}
\definecolor{hlblue}{HTML}{B5D9F6}
\definecolor{hllightgray}{HTML}{F8F9FA}
\definecolor{hlgray}{HTML}{E6E6E6}
\definecolor{hlcaseblue}{HTML}{00FFFF}
\definecolor{hlcasegreen}{HTML}{00FF00}
\definecolor{hlcaseyellow}{HTML}{FFFF00}
\definecolor{darkgreen}{HTML}{006400}
\definecolor{lightred}{RGB}{255, 230, 230}
\newcommand{\hlc}[2][hlgreen]{\begingroup\sethlcolor{#1}\hl{#2}\endgroup}
\newcommand{\improve}[1]{$~_{\textcolor{red}{+ #1}}$}

\newcommand{\difference}[1]{$~_{\textcolor{gray}{+ #1}}$}
\newcommand{\hlnew}[1]{#1}
\newcommand{\newtext}[1]{#1}

\newcommand{\name}{CARE}
\newcommand{\nameb}{CARE-Flow}
\newcommand{\namec}{CARE-Coord}

\title{CARE: Towards Clinical Accountability \\in Multi-Modal Medical Reasoning with \\an Evidence-Grounded Agentic Framework}

\author{
Yuexi~Du$^{1,2}$\thanks{Work done during internship at Microsoft Research Asia.},\quad
Jinglu~Wang$^{1}$\thanks{Corresponding Author.},\quad 
Shujie~Liu$^{1}$,\quad
Nicha~C.~Dvornek$^{2,3}$,\quad
Yan~Lu$^{1}$\\
\textsuperscript{1}Microsoft Research Asia\\
\textsuperscript{2}Department of Biomedical Engineering, Yale University\\
\textsuperscript{3}Department of Radiology \& Biomedical Imaging, Yale University\\
{\tt\small\{yuexi.du,nicha.dvornek\}@yale.edu,~\{jinglwa,yanlu\}@microsoft.com}
}

\iclrfinalcopy 
\begin{document}

\maketitle

\begin{abstract}
Large visual language models (VLMs) have shown strong multi-modal medical reasoning ability, but most operate as end-to-end black boxes, diverging from clinicians’ evidence-based, staged workflows and hindering clinical accountability. Complementary to this, expert visual grounding models can accurately localize regions of interest (ROIs), providing explicit, reliable evidence that improves both reasoning accuracy and trust.
In this paper, we introduce \textbf{\name}, advancing \textbf{C}linical \textbf{A}ccountability in multi-modal medical \textbf{R}easoning with an \textbf{E}vidence-grounded agentic framework. Unlike existing approaches that couple grounding and reasoning within a single generalist model, \name~decomposes the task into coordinated submodules to reduce shortcut learning and hallucination: a compact VLM proposes relevant medical entities; an expert entity-referring segmentation model produces pixel-level ROI evidence; and a grounded VLM reasons over the full image augmented by ROI hints. The VLMs are optimized with reinforcement learning with verifiable rewards to align answers with supporting evidence. Furthermore, a VLM coordinator plans tool invocation and reviews evidence-answer consistency, providing agentic control and final verification.
Evaluated on standard medical VQA benchmarks, our \textbf{\nameb}~(coordinator-free) improves average accuracy by \textbf{10.9\%} over the same size (10B) state-of-the-art (SOTA). With dynamic planning and reviewing, our \textbf{\namec}~yields a further gain, outperforming the heavily trained SOTA by \textbf{5.2\%}. 
Our experiments demonstrate that an agentic framework that emulates clinical workflows, incorporating decoupled specialists and explicit evidence, yields more accurate and accountable medical AI. Project page: \url{https://xypb.github.io/CARE-Project-Page/}.
\end{abstract}

\section{Introduction}
\label{sec:intro}

\begin{figure}[!t]
    \centering
    \includegraphics[width=\columnwidth]{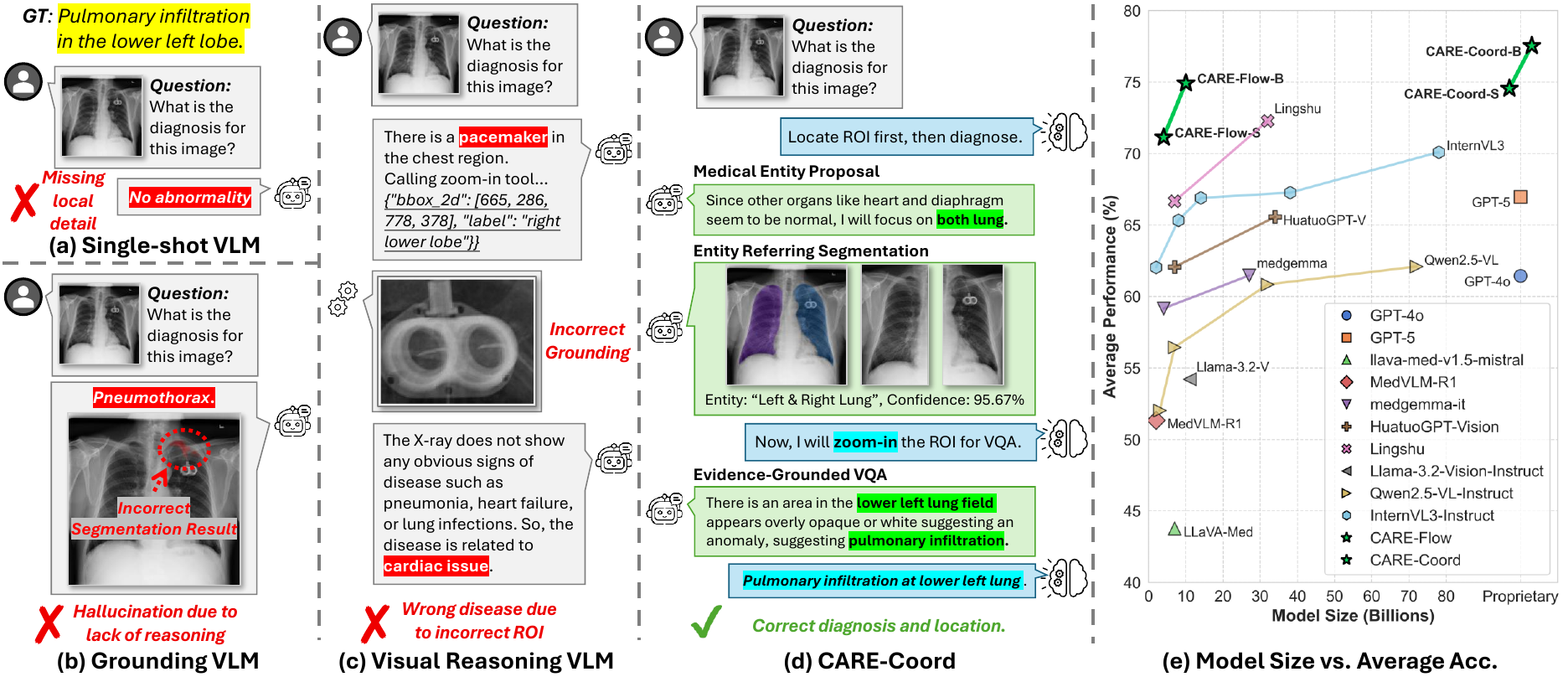}
    \vspace{-4mm}
    \caption{{\bf VLMs for medical reasoning.} (a) Single-shot VLMs often miss local evidence. (b) Grounding VLMs do not explicitly utilize ROI in reasoning. (c) Generalist visual reasoning VLMs fail with incorrect initial focus. (d) Our agentic \namec~performs grounded evidence-based reasoning and expert discussion, improving accountability. (e) Comparison of average medical VQA accuracy vs.\ model size. Models with unknown size appear in the rightmost panel.}
    \vspace{-8mm}
    \label{fig:teaser}
\end{figure}

Recent advances in visual language models (VLMs) have delivered strong results in medical image understanding and diagnostic visual question answering (VQA)~\citep{he2024foundation,xu2025lingshu,sellergren2025medgemma}. However, most current methods~\citep{dong2025seeing,hou2024vision,li2025core} adopt a monolithic, single-shot formulation that maps images and text directly to answers without explicitly localizing or verifying the supporting visual findings. This design invites shortcut learning and hallucination, especially under distribution shift, as fine-grained, case-relevant evidence is neither retrieved nor required, as illustrated in \cref{fig:teaser} (a). Unlike human clinical workflow, which localizes abnormalities, examines them at appropriate scales, and then decides based on explicit image evidence, such black-box inference undermines clinical reliability and accountability.

In response, some works augment VLMs with visual grounding~\citep{wu2025unibiomed,huang2025towards,luo2024vividmed,zhu2025guiding}, but typically treat grounding as an isolated perception head whose outputs are not fed back into reasoning for full use (\cref{fig:teaser} (b)). 
In general-domain VQA, concurrent works~\citep{fan2025grit,zhang2023towards,qi2024cogcom} interleave external image manipulations, \eg, zoom-in, crop, OCR, between grounding and reasoning to supply regions of interest (ROIs) to the chain of thought.
However, these methods couple perception and reasoning inside a single generalist model.
This coupling demands high-quality paired ROI-grounding and VQA supervision data and often costly multi-turn reinforcement learning (RL) to stabilize tool use~\citep{yang2025visionthink,zheng2025deepeyes}. 
Both tasks can degrade when such data are scarce. Compared with specialist visual grounding models \citep{liu2023grounding,ren2024grounded}, VLM-based grounding frequently misses tiny but clinically salient findings, weakening downstream reasoning.
Moreover, chaining all steps inside one model amplifies error propagation: early grounding errors bias subsequent reasoning and yield confident hallucinations, as illustrated in \cref{fig:teaser} (c). These limitations motivate an agentic framework that coordinates well-trained specialist tools and feeds grounded evidence back into reasoning.

To advance \textbf{C}linical \textbf{A}ccountability in multi-modal medical \textbf{R}easoning, we introduce an \textbf{E}vidence-grounded agentic framework, \textbf{\name}. 
As presented in \cref{fig:teaser} (d), given a user query and a medical image, \name~explicitly models the clinical diagnostic workflow to perform VQA in three stages: 
(1) Medical entity proposal: a question-conditioned VLM proposes candidate medical entities (\eg, anatomical structures and findings), which are fine-tuned via RL with verifiable reward (RLVR) to improve performance and accountability;
(2) Entity referring segmentation: given proposed entities, a tailored referring-segmentation model localizes the corresponding ROIs, producing pixel-level evidence;
(3) Evidence-Grounded VQA (EG-VQA): an EG-VQA model reasons over the full image and one of three evidence views commonly used in medical imaging: (i) a zoom-in crop for local detail, (ii) a binary mask for positional/spatial priors, or (iii) a global indicator when local evidence is unnecessary.
To operationalize agentic control, we introduce a dynamic coordinator \textbf{\namec}~that plans the tool-invocations, selects the most informative evidence view, and performs iterative answer review, mitigating hallucinations. For a coordinator-free, self-contained setting, \textbf{\nameb}~executes all three evidence views and aggregates EG-VQA outputs via simple rules.

We evaluated \name~on four standard medical VQA benchmarks~\citep{hu2024omnimedvqa,liu2021slake,ben2019vqa,lau2018dataset} spanning over ten image modalities and multiple organs, with results summarized in \cref{fig:teaser} (e). Extensive experiments validate the effectiveness of our clinician-inspired framework. 
\nameb~(totaling 10B parameters) shows strong competitive results on multiple benchmarks, outperforming comparable generalist models and demonstrating strong parameter efficiency.
demonstrating strong parameter efficiency. Adding the agentic coordinator, \namec, further improves performance by a large margin, showcasing the potential of agentic reasoning. Our contribution is summarized below:
\begin{compactitem}
    \item We propose \textbf{\name}, the first medical agentic framework for accountable medical visual reasoning. A dynamic coordinator plans tool use and conducts iterative answer review, reducing hallucinations via explicit evidence checks.
    \item We design a region-grounded reasoning workflow that feeds reliable, pixel-level evidence (referring segmentation, zoom-in views, or global indicators) back into VQA, improving both accuracy and accountability via accurate entity proposal and segmentation.
    \item Empirically, our \nameb~surpasses the same size (10B) SOTA baseline by \textbf{10.9\%}, and exceeds the heavily-trained SOTA baseline (Lingshu-32B~\citep{xu2025lingshu}) by \textbf{2.26\%}; \namec~further outperforms Lingshu-32B by \textbf{5.2\%}.
\end{compactitem}

\section{Related Work} 
\label{sec:related_works}

\textbf{Medical Multimodal Large Language Models.} General-purpose VLMs~\citep{bai2025qwen2,zhu2025internvl3,hurst2024gpt,meta2024llama,openai_gpt5_system_card_2025} lack expert medical knowledge. Early medical VLMs~\citep{li2023llava,moor2023med} used low-quality data~\citep{zhang2023biomedclip}, and while recent systems~\citep{chen2024huatuogpt,xu2025lingshu,lin2025healthgpt, he2024foundation} leverage better data and RL~\citep{lai2025med,pan2025medvlm}, most remain single-shot black boxes prone to hallucination. Agentic pipelines~\citep{zhu2025medagentboard,tang2023medagents,xia2025mmedagent,kim2024mdagents} or tool-use~\citep{li2024mmedagent,wang2025medagent,nath2025vila,he2025medorch} are also explored but typically lack the visual evidence needed for diagnostic reliability. Instead, our method, with visual-evidence supported reasoning, provides much better accountability for the answer.

\textbf{Grounded VLMs.} Research on grounded VLMs~\citep{zhang2024omg,rasheed2024glamm,yuan2025sa2va,zhang2024groundhog,lai2024lisa} and medical VLMs~\citep{huang2025towards,wu2025unibiomed,lin2025healthgpt,luo2024vividmed,chen2025mimo,huang2024refer,huang2025medseg} has focused on a grounded output, but typically treats grounding as an auxiliary multi-task optimization rather than using grounded clues to improve answer quality. These methods also require large-scale, fine-grained annotated data for supervised fine-tuning (SFT). In contrast, our method treats grounded visual evidence as supporting evidence for downstream reasoning, training a VLM specialized to utilize local visual clues, which naturally leads to a higher performance at testing~\citep{chen2024r}.

\textbf{Vision-Language Reasoning.} Since \citet{wei2022chain} proposed Chain-of-Thought (CoT), various methods~\citep{wei2022chain,yao2023tree,schulman2017proximal,rafailov2023direct} and RLVR~\citep{shao2024deepseekmath,yu2025dapo} have advanced reasoning. However, vision-language reasoning methods~\citep{yang2025visionthink,zheng2025deepeyes,fan2025grit,zhong2025focus,zhang2023towards,li2024vocot,yang2025look,qi2024cogcom} that focus on image content are often computationally expensive (e.g., multi-turn)~\citep{zheng2025deepeyes,fan2025grit,zhong2025focus,yang2025look} or require high-quality human-annotated data~\citep{fan2025grit,qi2024cogcom,li2024vocot,zhang2023towards}, limiting medical adoption. In contrast, our method uses model-proposed visual clues as direct inputs during both training and inference.

\section{Method}
\label{sec:method}

To advance multimodal medical reasoning while mitigating shortcut learning and hallucination, we decompose it into specialized sub-tasks and integrate expert visual tools with agentic coordination, aligning the pipeline with clinical practice and improving accountability with high-quality visual evidence and staged reasoning. 

\begin{figure}[!t]
    \centering
    \includegraphics[width=0.95\columnwidth]{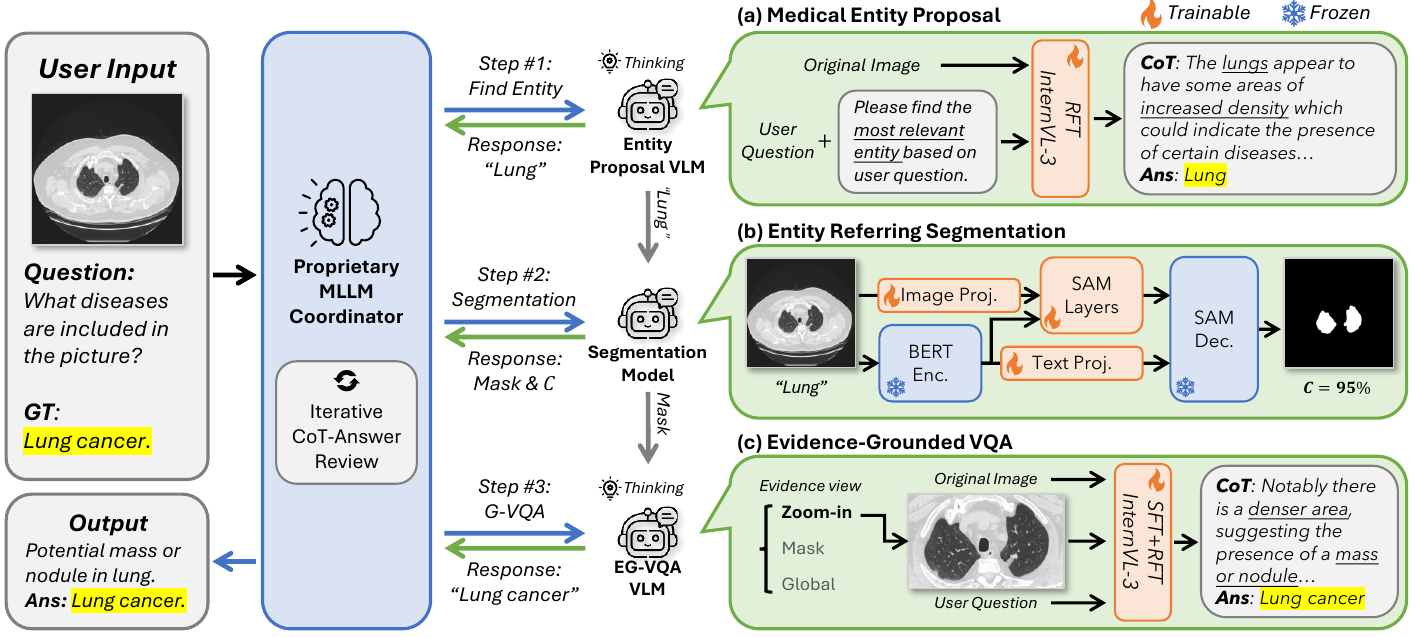}
    \vspace{-3mm}
    \caption{\textbf{Method overview.} The proposed \name~comprises a VLM coordinator and a set of task-specific expert models. The coordinator plans tool use and conducts answer review, invoking specialist models as needed. The expert set includes: (1) a question-conditioned entity-proposal VLM that identifies relevant anatomical structures/findings; (2) a referring segmentation model that localizes entities with pixel-level ROI evidence; and (3) an evidence-grounded VQA VLM that reasons over the image augmented with selected visual evidence (zoom-in, mask, or global indicator).
    }
    \vspace{-3mm}
    \label{fig:method}
\end{figure}

\textbf{Method Overview.} We detail the \name~framework with an overview in \cref{fig:method}. \name~takes a user question and a medical image, and executes three decomposed sub-tasks:
(1) Medical entity proposal. A question-prompted, compact VLM proposes relevant anatomical structures or findings. The VLM is fine-tuned with RLVR for evidence-consistent proposals.
(2) Entity-referring segmentation. A tailored referring-segmentation model localizes the entities and produces high-quality pixel-level evidence (ROI masks/regions).
(3) Evidence-Grounded VQA. A finetuned VQA model reasons over the full image augmented by the grounded evidence. 
We further introduce the agentic control with the coordinator \namec, which plans the tool invocation sequence, selects the most informative evidence view, and performs iterative chain-of-thought review before finalization. For a coordinator-free setting, \nameb~follows a static workflow that executes all evidence views and aggregates EG-VQA outputs by simple rules (\eg, majority vote).

\subsection{Medical Entity Proposal}
\label{sec:method_roi_proposal}

\textbf{Medical Entity.}
We first propose the most relevant entities in the image conditioned on the user’s question, mirroring a clinician’s workflow of hypothesizing which anatomical structures, findings, or devices are implicated. We refer to these as \textit{medical entities}. A compact VLM is fine-tuned with RLVR to prioritize proposals that support evidence-consistent answers. As no public dataset exists for this task, we synthesize training data: for each image, we randomly sample a segmentation mask/medical entity and generate a corresponding question, yielding paired (image, question, entity/mask) examples for supervision (see \cref{sec:app_data_synth} in detail).

\textbf{Reinforcement Fine-tuning (RFT) for entity proposal.}
We fine-tune the entity proposal VLM with RLVR. Instead of a binary accuracy reward, we use an embedding-similarity reward to capture semantic matches. For an input image-question pair, the model outputs a set of $P$ entity names $\hat{\mathcal{E}}=\{\hat{e}_i\}^{P}_{i=1}$. With $Q$ ground truth $\mathcal{E}=\{e_i\}^Q_{i=1}$, a small embedding language model $\mathrm{SLM}$~\citep{wang2020minilm} computes pairwise cosine similarities $s_{i,j}=\mathrm{sim}\big(\mathrm{SLM}(\hat{e}_i),\mathrm{SLM}(e_j)\big)$, forming a matrix $S\in\mathbb{R}^{P\times Q}$. We apply the Kuhn–Munkres algorithm~\citep{kuhn1955hungarian} to find an optimal bijection $\mathcal{K}=\{(\hat{e}_i,e_j)\}^{\min(P,Q)}$ maximizing total similarity, and define the similarity reward as:
\begin{equation}
    R_{\mathrm{sim}}(\hat{\mathcal{E}},\mathcal{E})
    = \frac{1}{\min(P,Q)} \sum_{(\hat{e}_i,e_j)\in \mathcal{K}} s_{i,j}.
    \label{eq:reward_sim}
\end{equation}
Using soft similarity reward not only avoids $0$-gradient issue during RFT but also helps mitigate the domain gap between synthetic data and real user questions. It is not forcing an exact match, but rather maximizing semantic similarity.
We further include a count reward that discourages empty or overly long proposals,
$R_{\mathrm{count}}(\hat{\mathcal{E}}) = 1$ if $0 < P \le 5$, and $0$ otherwise;
and a repetition penalty $R_{\mathrm{repetition}}=\tfrac{1}{r+1}$ with $r$ the number of repeated entities. The total reward takes the form:
\begin{equation}
    R_{\mathrm{Entity}} \;=\; R_{\mathrm{sim}} + R_{\mathrm{count}} + R_{\mathrm{repetition}} + R_{\mathrm{format}},
    \label{eq:roi_reward}
\end{equation}
where $R_{\mathrm{format}}$ enforces the \texttt{<think>} and \texttt{<answer>} tags to wrap CoT and answer respectively. The resulting VLM proposes entities closely aligned with the user’s query, mirroring the clinician’s first “where to look” step. The proposed entities serve as inputs to the segmentation model.

\subsection{Entity Referring Segmentation}
\label{sec:method_segment}

We build an entity referring segmentation model based on SA-Med-2D~\citep{cheng2023sammed2d} as shown in \cref{fig:method} (b). Given a pre-trained SAM-style segmenter with image projector $ \mathrm{Proj}_I $, SAM encoder layers $ \mathrm{Enc}_{SAM} $, and SAM mask decoder $ \mathrm{Dec}_M $, we augment it with a frozen, lightweight BERT-style biomedical text encoder~\citep{alsentzer2019publicly} $ \mathrm{Enc}_T $ and an embedding projector $ \mathrm{Proj}_T $. For an input image–entity pair $(x^I,\hat{e})$, where $e\in\mathcal{\hat{E}}$, we encode the image and entity into token sequences $t_I=\mathrm{Proj}_I(x^I)$ and $t_T=\mathrm{Enc}_T(\hat{e})$, respectively. We then concatenate them with binary modality token embeddings $t_{mod}$ to form the SAM encoder input $t=\operatorname{concat}(t_I,t_T)+t_{mod}$. Inspired by positional encodings, $t_{mod}$ is set to 0 for image tokens and 1 for text tokens. We only use the image tokens from the output, \ie, 
first $|t_I|$ tokens, as key and value for $ \mathrm{Dec}_M $. Meanwhile, we project $t_T$ with the $ \mathrm{Proj}_T $ and use it as the query. The final segmentation mask is given by:
\begin{equation}
    M=\mathrm{Dec}_M\!\big(\mathrm{Enc}_{SAM}(t)[0:|t_I|],\, \mathrm{Proj}_T(t_T)\big).
    \label{eq:seg}
\end{equation}
During fine-tuning, we only update $\mathrm{Proj}_I$, $\mathrm{Enc}_I$, and $\mathrm{Proj}_T$, thereby equipping the pre-trained medical SAM model with the ability of referring segmentation. Because masks may be imperfect, we compute a confidence score $C$ from the mask probability map $M_p$ via $C(M_p) \;=\; 1 - \tfrac{\operatorname{Entropy}(M_p)}{\log(2)}$,
and pass $C$ downstream so the coordinator and EG-VQA model can filter low-quality segmentations.

\subsection{Evidence-Grounded Visual Question Answering}
\label{sec:method_gvqa}

\textbf{Evidence-grounded VQA.}
We treat the segmentation mask as an \emph{additional hint} and design three visual evidence types that reflect clinical practice and avoid information loss: (1) \textbf{Zoom-in:} we zoom in and crop around the ROI to provide a detailed, higher-resolution local view; (2) \textbf{Mask:} we feed the binary mask as a separate signal that acts as an attention-amplification prior highlighting positional and spatial context; (3) \textbf{Global:} we provide an all-ones mask when no segmentation is available or when the task depends on global context (\eg, modality or imaging axis). This scheme allows the EG-VQA model to adapt to different question types while remaining efficient. We drop input masks whose confidence falls below a confidence threshold $\tau_C$ empirically chosen based on \cref{sec:app_add_exp}, preventing low-quality segmentation from harming decisions; in the worst case, the VLM falls back to its ROI-free behavior. We append mask metadata in the prompt as \texttt{"<image> (instance: \{NAME\}, confidence: \{CONFIDENCE\}\%)"}. The clue is concatenated after the input image to form a multi-image input. We choose not to directly overlay the mask over the original medical image since the pixel value and image contrast may have physical meaning.

\textbf{Fine-tuning the EG-VQA VLM.}
We fine-tune the EG-VQA VLM in a two-stage manner. First, we use the trained entity proposal VLM and the referring segmentation model to annotate raw VQA datasets with visual clues. Second, we perform SFT followed by RFT on the combined data, including all three clue types. For RFT we add a CoT-length reward $R_{\mathrm{length}}(\hat{y})=0.25 \cdot \min\!\big(1, \tfrac{|\hat{y}|}{L}\big)$, where $\hat{y}$ is the generated reasoning and $L$ is a preset maximum reasoning length, alongside a binary accuracy reward $R_{\mathrm{acc}}$ (no external LLM verifier) and a format reward $R_{\mathrm{format}}$ following \cref{sec:method_roi_proposal}. The final reward for EG-VQA model is:
\begin{equation}
    R_{\mathrm{EG\text{-}VQA}} = R_{\mathrm{acc}} + R_{\mathrm{format}} + R_{\mathrm{length}}.
    \label{eq:gvqa_reward}
\end{equation}
The fine-tuned EG-VQA model is capable of handling three different types of visual clues and making more accurate evidence-supported decisions.

\subsection{Reinforcement Learning with Verifiable Reward}
\label{sec:method_prelim}

We fine-tune expert VLMs using RLVR to improve the answer accountability and generalizability. We specifically employ the DAPO~\citep{yu2025dapo} algorithm. With outputs $\{y_i\}^G_{i=1}$ generated by reference model $\pi_{\mathrm{ref}}$ for input $x$, we update the policy model $\pi_\theta$ using the following objective:
\begin{equation}
    \mathcal{J}_{\mathrm{DAPO}}(\theta)
    = \mathbb{E}_{y_i \sim \pi_{\mathrm{ref}}(\cdot \mid x)}\!\left[
      \frac{1}{\sum_{i=1}^{G} |y_i|}
      \sum_{i=1}^{G}\sum_{j=1}^{|y_i|}
      \min\!\Big( r_{i,j} A_{i,j},
      \operatorname{clip}\!\big(r_{i,j}, 1-\epsilon_{l}, 1+\epsilon_{h}\big) \Big)
    \right],
    \label{eq:dapo}
\end{equation}
where $r_{i,j}=\frac{\pi_{\theta}(y_{i,j}\mid x_i,y_{i,<j})}{\pi_{\mathrm{ref}}(y_{i,j}\mid x_i,y_{i,<j})}$, and the advantage is group-normalized as $A_{i,j}=\frac{R_i-\operatorname{mean}(\{R_i\}_{i=1}^{G})}{\operatorname{std}(\{R_i\}_{i=1}^{G})}$. $\epsilon_{l}$ and $\epsilon_{h}$ are the upper and lower clip thresholds. More detail is in \cref{sec:app_rlvr}. Recent studies~\citep{chu2025sft, ma2025learning} suggest SFT injects new knowledge (memorization), while RFT improves existing capabilities by adjusting output to generate a reasonable Chain-of-Thought (CoT). We choose to apply RFT to each expert VLM to improve accountability via CoT under limited data.

\subsection{Coordinating Expert Models for Vision Reasoning}
\label{sec:method_magnify}

We propose {\name} in two modes that mimic clinical workflows: a static pipeline and a dynamic, coordinator-driven agent. Instead of a single model, we decouple the reasoning process into collaborating expert models. This approach optimizes each model for its specific task and uses independent reasoning with post-verification to prevent the amplification of errors from prior steps.

\textbf{Static workflow.}
Our static workflow framework, \textbf{\nameb}, processes an image-question pair $(x_I,x_T)$ through a sequential pipeline: \texttt{Entity Proposal VLM} $\rightarrow$ \texttt{Segmentation Model} $\rightarrow$ \texttt{EG-VQA VLM}. The entity proposal model outputs a set of $P$ entities $\hat{\mathcal{E}}=\{e_i\}^P_{i=1}$. The segmentation model then produces corresponding masks $\mathcal{M}=\{M_i\}^P_{i=1}$ and confidences $\mathcal{C}=\{C\}^P_{i=1}$, and we discard masks where $C_i < \tau_C$. Lacking a coordinator to select the best visual clue, we call the EG-VQA model three times, once for each clue type, and use majority vote to decide the final answer.

\begin{table}[t]
\centering
\caption{\textbf{Quantitative results on medical VQA benchmarks.} We report medical VQA accuracy (\%) on four standard benchmarks: OMVQA-3k~\citep{hu2024omnimedvqa}, VQA-RAD~\citep{lau2018dataset}, SLAKE~\citep{liu2021slake} and VQA-Med-2019~\citep{ben2019vqa}. Open-ended questions are scored by GPT-4o against ground-truth answers. Our segmentation model is smaller than 1B.
We highlight medical expert VLMs in \hlc[hlgray]{gray} and ours in \hlc[hlgreen]{green}.
}
\label{tab:main_res}
\vspace{1mm}
\resizebox{0.90\textwidth}{!}
{

\begin{tabular}{lccccc}
\toprule

\multicolumn{1}{c}{\textbf{Model}} & \textbf{OMVQA-3k} & \textbf{VQA-RAD} & \textbf{SLAKE} & \textbf{VQA-Med-2019} &  \textbf{Overall} \\ \midrule\midrule
\multicolumn{6}{c}{\textit{Proprietary}} \\ \midrule
GPT-4o~\citep{hurst2024gpt} & 64.07 & 58.54 & 63.55 & 59.60 & 61.44 \\
GPT-5~\citep{openai_gpt5_system_card_2025} & 74.73 & 63.19 & 67.75 & 62.20 & 66.97 \\ \midrule
\multicolumn{6}{c}{\textit{Open-source}} \\ \midrule
Llama-3.2-11B-Vision~\citep{meta2024llama} & 43.10 & 53.22 & 63.17 & 57.40 & 54.22 \\
Qwen2.5-VL-7B~\citep{bai2025qwen2} & 61.40 & 54.10 & 59.73 & 50.60 & 56.46 \\
Qwen2.5-VL-32B~\citep{bai2025qwen2} &
65.10 & 61.20 & 65.46 & 51.60 & 60.84 \\
InternVL3-8B~\citep{zhu2025internvl3} & 75.97 & 61.86 & 66.13 & 57.40 & 65.34 \\
InternVL3-38B~\citep{zhu2025internvl3} & 78.57 & 62.97 & 68.70 & 58.80 & 67.26 \\
DeepEyes-7B~\citep{zheng2025deepeyes} & 57.40 & 56.10 & 61.16 & 52.20 & 56.72 \\

\rowcolor[HTML]{E6E6E6} llava-med-v1.5-mistral-7b~\citep{li2023llava} & 45.30 & 41.91 & 50.86 & 37.00 & 43.77 \\
\rowcolor[HTML]{E6E6E6} MedVLM-R1-2B~\citep{pan2025medvlm} & 72.07 & 41.46 & 46.47 & 45.40 & 51.35 \\
\rowcolor[HTML]{E6E6E6} medgemma-4b~\citep{sellergren2025medgemma} & 61.50 & 58.09 & 69.66 & 47.40 & 59.16 \\
\rowcolor[HTML]{E6E6E6} medgemma-27b~\citep{sellergren2025medgemma} & 64.23 & 62.75 & 70.52 & 48.40 & 61.47 \\
\rowcolor[HTML]{E6E6E6} HuatuoGPT-Vision-7B~\citep{chen2024huatuogpt} & 70.70 & 59.87 & 60.50 & 57.20 & 62.07 \\
\rowcolor[HTML]{E6E6E6} HuatuoGPT-Vision-34B~\citep{chen2024huatuogpt} & 76.80 & 60.75 & 64.12 & \uline{60.60} & 65.57 \\
\rowcolor[HTML]{E6E6E6} Lingshu-7B~\citep{xu2025lingshu} & 73.17 & 58.54 & 76.15 & 58.80 & 66.66 \\
\rowcolor[HTML]{E6E6E6} Lingshu-32B~\citep{xu2025lingshu} & 83.97 & \uline{64.75} & 82.25 & 58.20 & 72.29 \\ \midrule

\rowcolor{hlgreen} \textbf{\nameb-S (4B)} & 94.53 & 56.32 & 78.44 & 53.60 & 70.72 \\
\rowcolor{hlgreen} \textbf{\namec-S} & \uline{97.70} & 62.75 & 77.19 & \uline{60.60} & 74.56 \\
\rowcolor{hlgreen} \textbf{\nameb-B (10B)} & 96.17 & 63.64 & \textbf{83.21} & 56.60 & \uline{74.91} \\
\rowcolor{hlgreen} \textbf{\namec-B} & \textbf{97.97} & \textbf{68.29} & \uline{83.11} & \textbf{60.80} & \textbf{77.54} \\ \bottomrule
\end{tabular}

}
\vspace{-6mm}
\end{table}

\textbf{Dynamic coordination.}
Our dynamic agent, \textbf{\namec}, employs a powerful VLM as the coordinator (\cref{fig:method}). The coordinator can \emph{plan} which models to call, \emph{act} using tool calls, and \emph{review} intermediate outputs to verify reasoning quality. This dynamic process improves decision quality and mitigates hallucinations. We instruct the coordinator to verify the VLMs' reasoning logic rather than making clinical judgments. The coordinator also improves efficiency by choosing the optimal visual clue, or even skipping localization entirely for global questions (e.g., about image modality), which reduces tool calls. Among the models we tested, GPT-5~\citep{openai_gpt5_system_card_2025} was the best-performing coordinator. We also detail experiments on training a small VLM for this task in~\cref{sec:app_rft_coordinator}.

\textbf{Iterative CoT-Answer Review.}
A small expert VLM may generate answers that are inconsistent with its reasoning. This is because our rule-based reward $R_{\mathrm{acc}}$ only verifies the final answer, not the preceding chain-of-thought. While adding an LLM verifier could fix this, it would be computationally expensive. Instead, we use the coordinator for \emph{iterative CoT-answer review} post inference. We instruct the coordinator to check the consistency of each thought-answer pair. If they disagree, the coordinator can re-run the expert model or correct the pair using its own reasoning.

\section{Experiments}
\label{sec:exps}

In this section, we evaluate the performance of \name~on four standard medical VQA benchmarks and compare it against state-of-the-art baselines. Our goal is to answer the following research questions. \textbf{RQ1}: \textit{Is \name~performing better than other reasoning or non-reasoning VLMs?} \textbf{RQ2}: \textit{Is it helpful to include visual evidence for the VQA tasks and how to better make use of it?} \textbf{RQ3}: \textit{Does including the coordinator improve the capability of \name?} \textbf{RQ4}: \textit{What is the influence of using the entity proposal and segmentation model?}

\textbf{Datasets.}
For the entity proposal model, we train it with \textbf{SA-Med-20M}~\citep{ye2023sa} dataset. We create a synthetic dataset of 10k training and 1k testing question-ROI pairs.
As for the segmentation model, we train it with 170k image-mask pairs from \textbf{SA-Med-20M}~\citep{ye2023sa} dataset. We evaluate segmentation performance on the \textbf{MeCo-G}~\citep{huang2025towards} dataset.
For the \textbf{VQA} task, we use \textbf{OmniMedVQA}~\citep{hu2024omnimedvqa}, \textbf{VQA-RAD}~\citep{lau2018dataset}, and \textbf{SLAKE}~\citep{liu2021slake} for \textbf{in-domain (ID)} training. All the ID data are combined during training. For OmniMedVQA, we randomly create a 4k/3k split for training/testing.
Besides the test set for ID data, we further use \textbf{VQA-Med-2019}~\citep{ben2019vqa} for \textbf{out-of-domain (OOD)} evaluation. Both open \& closed-ended questions are included. For open-ended questions, we use GPT-4o~\citep{hurst2024gpt} to judge the accuracy of the answers.

\begin{table}[t]

\vspace{-3mm}
\centering
\caption{\textbf{Ablation on grounded VQA.} 
We ablate the 8B EG-VQA components during training, varying training visual evidence and coordinator configurations. Only one type of visual evidence is used for inference.
\nameb~and \namec~ are highlighted in \hlc[hlblue]{blue} and \hlc[hlgreen]{green}, respectively.
}
\label{tab:ablate_vqa}
\vspace{1mm}
\resizebox{0.87\textwidth}{!}
{

\begin{tabular}{ccccccccccc}
\toprule
\multicolumn{3}{c}{\multirow{2}{*}{\textbf{Training Visual Clue}}} & \multicolumn{2}{c}{\multirow{2}{*}{\textbf{Coordinator}}} & \multicolumn{6}{c}{\textbf{Datasets}} \\ \cmidrule(l){6-11} 
\multicolumn{3}{c}{} & \multicolumn{2}{c}{} & \multicolumn{4}{c}{\textbf{ID}} & \textbf{OOD} & \multirow{2}{*}{\textbf{Overall}} \\ \cmidrule(r){1-3} \cmidrule(r){4-5} \cmidrule(r){6-9} \cmidrule(r){10-10}
\textbf{Mask} & \textbf{Zoom} & \textbf{Global} & \textbf{Planning} & \textbf{Review} & \textbf{OMVQA} & \textbf{VQA-RAD} & \textbf{SLAKE} & \textbf{Avg.} & \textbf{VQA-Med-2019} &  \\ \midrule
 &  &  &  &  & 94.5 & 60.5 & 78.8 & 77.9 & 56.0 & 72.4\difference{0.0} \\
\cmark &  &  &  &  & 95.4 & 61.6 & 81.8 & 79.6 & 54.0 & 73.2\improve{0.8} \\
 & \cmark &  &  &  & 95.1 & 61.2 & 82.3 & 79.5 & 56.8 & 73.8\improve{1.4} \\
\cmark & \cmark &  &  &  & 95.6 & 62.0 & 83.1 & 80.2 & 55.6 & 74.1\improve{1.7} \\
\rowcolor{hlblue} 
\cmark & \cmark & \cmark &  &  & 96.1 & 63.6 & \textbf{83.2} & 81.0 & 56.6 & 74.9\improve{2.5} \\
\cmark & \cmark & \cmark & \cmark &  & 95.9 & 65.1 & 81.3 & 80.8 & 53.4 & 74.8\improve{2.4} \\
\rowcolor{hlgreen} 
\cmark & \cmark & \cmark & \cmark & \cmark & \textbf{97.9} & \textbf{68.2} & 83.1 & \textbf{83.1} & \textbf{60.8} & \textbf{77.5}\improve{\bm{5.1}} \\ \bottomrule
\end{tabular}

}
\vspace{-7mm}
\end{table}
\begin{table}[t]
  \centering
  \begin{minipage}[t]{0.52\textwidth}
    \centering
    \caption{\textbf{Ablation on training strategy for EG-VQA.} 
    We ablate different training strategies for EG-VQA VLM. 
    We adopt the \nameb~in this ablation to exclude the coordinator's effects.
    }
    \label{tab:training}
    \vspace{1mm}
    \resizebox{\linewidth}{!}
    {
    \begin{tabular}{lccc}
    \toprule
    \multicolumn{1}{c}{\textbf{Method}} & \textbf{ID} & \textbf{OOD} & \textbf{Overall} \\ \midrule \midrule
    - & 67.9 & 57.4 & 65.3\difference{0.0} \\
    ~+ SFT & 77.8 & 56.6 & 72.5\improve{7.2} \\
    ~+ GRPO & 75.2 & 54.0 & 69.9\improve{4.6} \\
    ~+ DAPO & 77.0 & 54.2 & 71.3\improve{6.0} \\
    ~+ SFT + DAPO & 79.3 & 56.2 & 73.5\improve{8.2} \\
    \rowcolor{hlblue}~+ SFT + DAPO + $R_{\mathrm{length}}$ (\nameb)& \textbf{81.0} & \textbf{56.6} & \textbf{74.9}\improve{\bm{9.6}} \\ \bottomrule
    \end{tabular}
    }
  \end{minipage}
  \hfill
  \begin{minipage}[t]{0.46\textwidth}
    \caption{\textbf{Ablation on coordinator.} 
    We ablate different coordinators. 
    ``\textbf{S}'' denotes using a single selected visual evidence. 
    }
    \label{tab:coordinator}
    \vspace{1mm}
    \centering
    \resizebox{\linewidth}{!}
    {
    \begin{tabular}{lcccc}
    \toprule
    \multicolumn{1}{c}{\textbf{Coordinator}} & \textbf{Infer. Clue} & \textbf{ID} & \textbf{OOD} & \textbf{Overall} \\ \midrule\midrule
    N/A & Mask & 80.8 & 56.8 & 74.8 \\
    N/A & Zoom-in & \uline{81.1} & 56.2 & 74.9 \\
    N/A & Global & 80.5 & 54.4 & 74.0 \\
    N/A & Mask \& Zoom & 80.9 & 55.0 & 74.4 \\
    \rowcolor{hlblue} Majority Vote (\nameb) & \textbf{S} & 81.0 & 56.6 & 74.9 \\ \midrule
    InternVL3-38B & \textbf{S} & 79.7 & 56.8 & 74.0 \\
    GPT-4o & \textbf{S} & 79.6 & 54.2 & 73.3 \\
    \rowcolor{hlgreen} 
    GPT-5 (\namec) & \textbf{S} & \textbf{83.1} & \textbf{60.8} & \textbf{77.5} \\ \midrule
    InternVL3-8B + RFT & \textbf{S} & 80.7 & \uline{58.2} & \uline{75.1} \\ \bottomrule
    \end{tabular}
    }
  \end{minipage}
  \vspace{-8mm}
\end{table}

\textbf{Implementation Details.}
The segmentation model is built on SA-Med-2D~\citep{cheng2023sammed2d}, which is much smaller (600M) than large VLMs; more details are in \cref{sec:app_implementation}. The entity proposal VLM is finetuned using InternVL3-2B~\citep{zhu2025internvl3} with similar rewards in MiniLM-L6-v2~\citep{wang2020minilm}. The EG-VQA VLM adopts InternVL3-2B/8B models, fine-tuned on the in-domain VQA datasets. 
We denote variants with the suffixes “-S/B,” using 2B/8B EG-VQA VLMs, respectively. With the 2B entity-proposal model and a relatively small segmentation module, CARE-Flow-S/B totals 4B/10B parameters.
The default coordinator (\namec) adopts GPT-5~\citep{openai_gpt5_system_card_2025}. 
We set the mask confidence threshold $\tau_C=70\%$ in segmentation following our ablation~\cref{sec:app_add_exp}. More details are available in~\cref{sec:app_implementation,,sec:app_gvqa_prompt,sec:app_entity_prompt,sec:app_prompt}.

\textbf{Baselines.}
We compare our model against baselines from several categories.
\textbf{Proprietary Models}: GPT-4o~\citep{hurst2024gpt} and GPT-5~\citep{openai_gpt5_system_card_2025}.
\textbf{General VLMs}: Llama-3.2 Vision~\citep{meta2024llama}, Qwen2.5-VL~\citep{bai2025qwen2}, InternVL3~\citep{zhu2025internvl3}, and the visual reasoning model DeepEyes~\citep{zheng2025deepeyes}.
\textbf{Medical VLMs}: LLaVA-Med~\citep{li2023llava}, medgemma~\citep{sellergren2025medgemma}, HuatuoGPT-Vision~\citep{chen2024huatuogpt}, Lingshu~\citep{xu2025lingshu}, and the reasoning model MedVLm-R1-2B~\citep{pan2025medvlm}. We also compare with reported values from~\citep{yu2025fine,cui2024biomedical,lin2023pmc} in \cref{tab:new_baseline}.
We also benchmark different \textbf{Segmentation Models}: RecLMIS~\citep{huang2024cross}, LISA~\citep{lai2024lisa}, MedPLIB~\citep{huang2025towards}, UniBiomed~\citep{wu2025unibiomed}, and BiomedParse~\citep{zhao2024biomedparse}. Since not all segmentation models accept text prompts, we only compare with BiomedParse in the VQA benchmarks.

\begin{table}[t]
  \centering
  \vspace{-4mm}
  \begin{minipage}[t]{0.445\textwidth}
    \centering
    \caption{\textbf{Ablation on segmentation model.} 
    We evaluate segmentation models on MeCo-G dataset, and their impact on medical VQA. 
    Note that only BiomedParse can be adapted to referring segmentation.
    Results with $^\ast$ are reported by \citet{huang2025towards}. 
    }
    \label{tab:segment}
    \vspace{1mm}
    \resizebox{\linewidth}{!}
    {
    \begin{tabular}{lccccc}
    \toprule
    \multirow{2}{*}{\textbf{\begin{tabular}[c]{@{}c@{}}Segmentation \\ Model\end{tabular}}} & \multicolumn{4}{c}{\textbf{MeCo-G Mean Dice Score}} & \multirow{2}{*}{\textbf{\begin{tabular}[c]{@{}c@{}}Overall\\ MedVQA\end{tabular}}} \\ \cmidrule(lr){2-5}
     & \textbf{Der} & \textbf{CT} & \textbf{PET} & \textbf{Avg.} &  \\ \midrule\midrule
    RecLMIS$^\ast$ & 88.8 & 74.9 & 81.1 & 81.6 & - \\
    LISA-7B$^\ast$ & 81.3 & 52.6 & 54.2 & 62.7 & - \\
    MedPLIB-7B$^\ast$ & 79.9 & 59.8 & 64.5 & 68.1 & - \\
    UniBiomed & 63.1 & 8.5 & 18.5 & 24.9 & - \\
    BiomedParse & 82.6 & 8.4 & 20.3 & 30.1 & 74.1 \\
    \rowcolor{hlgreen} \textbf{Ours} & \textbf{92.5} & \textbf{75.6} & \textbf{82.5} & \textbf{81.9} & \textbf{77.5} \\ \bottomrule
    \end{tabular}
    }
  \end{minipage}
  \hfill
  \begin{minipage}[t]{0.525\textwidth}
    \caption{\textbf{\hlnew{Ablation on entity-proposal VLM.}} 
    We ablate training strategies for the entity-proposal VLM and measure their impact on entity proposal, segmentation, and medical VQA. 
    Baseline\#1/2/3 are of our framework using the 2B entity proposal VLM trained with different matching and rewards. ``G'' and ``KM'' are greedy and KM-based matching, respectively.
   }
    \label{tab:roi_prop}
    \vspace{1mm}
    \centering
    \resizebox{\linewidth}{!}
    {
    \begin{tabular}{lccccc}
    \toprule
    \textbf{\begin{tabular}[c]{@{}l@{}}Proposal \\ Model\end{tabular}} & \textbf{\begin{tabular}[c]{@{}c@{}}Entity\\ Match\end{tabular}} & \textbf{Reward} & \textbf{\begin{tabular}[c]{@{}c@{}}Entity\\ Accuracy\end{tabular}} & \textbf{\begin{tabular}[c]{@{}c@{}}Mask\\ Dice\end{tabular}} & \textbf{\begin{tabular}[c]{@{}c@{}}Overall\\ MedVQA\end{tabular}} \\ \midrule
    GPT-5 & - & - & 40.5 & 50.5 & 72.7 \\
    Baseline\#1 & G & Bin. & 72.8 & 41.0 & 75.1 \\
    Baseline\#2 & G & Sim. & 71.7 & 37.0 & 74.2 \\
    Baseline\#3 & KM & Bin. & 74.1 & 53.9 & 75.7 \\
    \rowcolor{hlgreen} \textbf{Ours} & KM & Sim. & \textbf{85.2} & \textbf{73.4} & \textbf{77.5} \\ \bottomrule
    \end{tabular}
    
    }
  \end{minipage}
  \vspace{-6mm}
\end{table}

\subsection{Main Results}
\label{sec:exps_main_results}

Evaluation in \cref{tab:main_res} demonstrates the superiority of our proposed method that mimics the clinician's diagnosis process, answering \textbf{RQ1}.
As shown in \cref{tab:main_res}, our constant 10B model (\nameb-B) achieves an SOTA average accuracy of 74.91\%, surpassing the much larger Lingshu-32B (72.29\%) by 2.6\%. This demonstrates our architecture's parameter efficiency, as even our 4B model (\nameb-S) outperforms models up to 38B parameters. While these general medical expert VLMs perform better than their base model due to vast domain-specific pre-training, our \name~surpassed them with limited fine-tuning data and computational resources. Comparing with DeepEyes-7B~\citep{zheng2025deepeyes}, a single-model vision reasoning VLM, our model shows a much more significant improvement against its baseline, demonstrating the effectiveness of our proposed multi-stage reasoning scheme. 
After switching to Agent mode, our \namec~gains a uniform performance improvement of $\sim3\%$ from its workflow version. This is not only coming from more reasoning, visual clue choosing, but also from the final answer reviewing by the coordinator. We further note that introducing a coordinator can help greatly improve the generalizability of OOD data, gaining over 6\% improvement on a small version of \name.
\namec~is not the best performing method on SLAKE since the dataset relies less on local visual clues, and its evaluation set is relatively small.

\begin{figure}[!t]
    \centering
    \includegraphics[width=0.95\columnwidth]{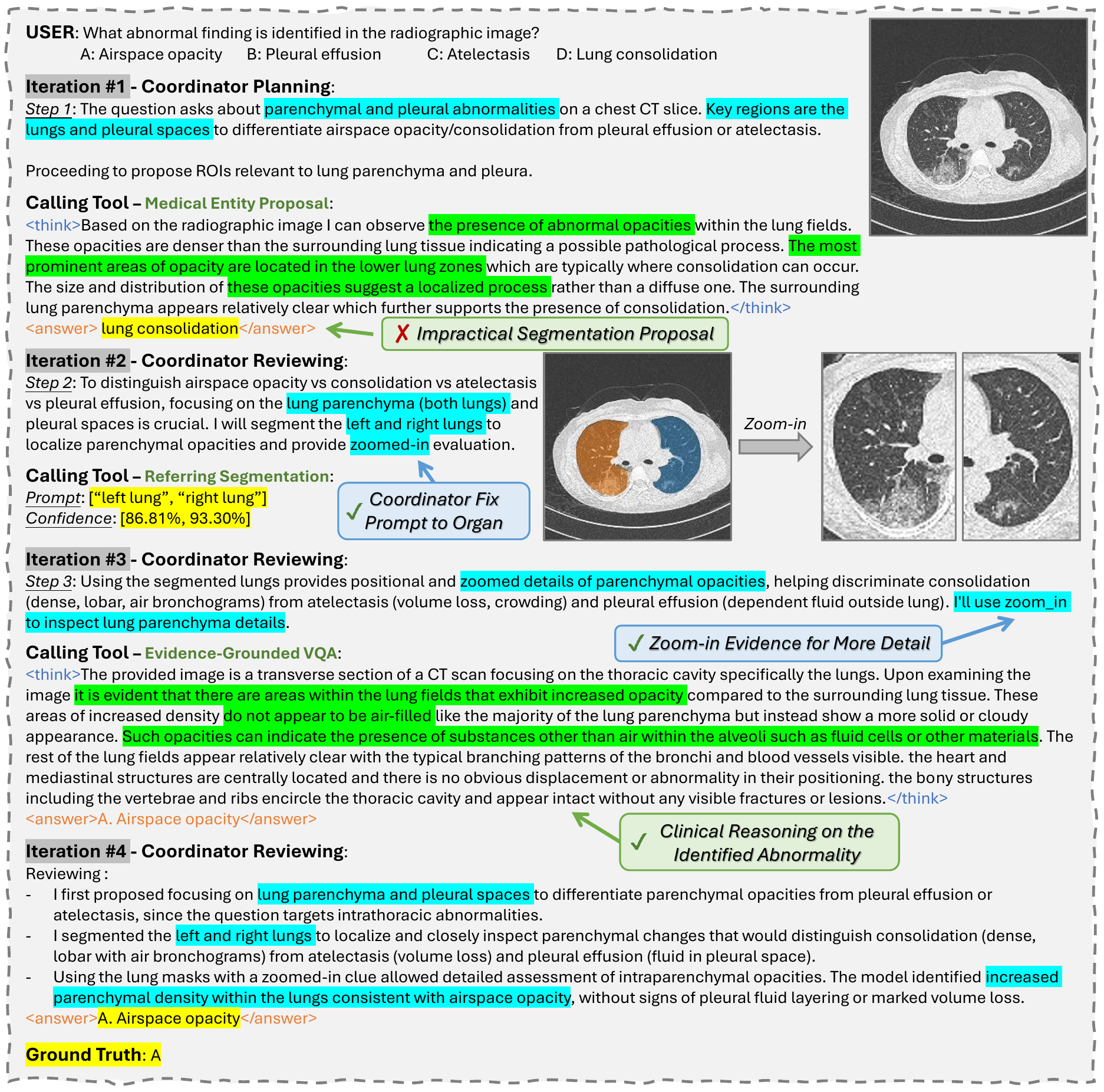}
    \vspace{-4mm}
    \caption{{\bf Case Study.} 
    We present the complete reasoning trace for a CT disease identification question.
    Key information from the coordinator is highlighted in \hlc[hlcaseblue]{blue}, model reasoning in \hlc[hlcasegreen]{green}, and each model’s final answer in \hlc[hlcaseyellow]{yellow}. 
    }
    \vspace{-6mm}
    \label{fig:case_study}
\end{figure}

\textbf{Qualitative Study.} We provide qualitative results in the \cref{fig:case_study} with a full reasoning process for a VQA pair in the test set. We note that our coordinator not only chose the correct visual clue type but also corrected an impractical proposal from entity proposal VLM, which helps the EG-VQA model to focus on the local detail. This highlights the effectiveness of the coordinator in planning and reviewing during inference time. We provide more examples that require more complex reasoning or clinically specific knowledge in \cref{sec:app_case}, \cref{fig:case_study1,fig:case_study2,fig:case_study3,fig:case_study4,fig:case_study5,fig:case_study6}.

\subsection{Ablation Study}
\label{sec:exps_ablation}

Without loss of generality, all ablations use the 2B entity proposal VLM and the 8B EG-VQA VLM.

\textbf{Ablation on EG-VQA Model.} Ablations in \cref{tab:ablate_vqa} confirm the effectiveness of visual clues (\textbf{RQ2}), as our full static model outperforms a no-clue baseline by 2.5\%. More diverse visual clues can improve the robustness at test time. Furthermore, a dynamic coordinator that also reviews the final answer (row 8) provides the largest benefit, improving accuracy to 77.5\%, while a coordinator without reviewing cannot guarantee a better result compared with \nameb.

\textbf{Ablation on Training Strategy.} We ablate different strategies in \cref{tab:training}. In general, using SFT or RFT individually only yields moderate improvement, while combining them results in $\sim1\%$ improvement. Meanwhile, we note that DAPO is generally better than GRPO. Lastly, we note that using an additional length reward also helps improve the final performance by $\sim1.4\%$ as it forces the model to provide longer reasoning and make use of visual evidence.

\textbf{Ablation on Coordinator.}
We evaluate the influence of using different coordinators in \cref{tab:coordinator}, where to note \namec~with robust coordinator beats all other variants, answering \textbf{RQ3}. Among all three evidence views, zoom-in stands out with the best performance. We also note that using both zoom-in and mask at inference may harm the performance, which may be influenced by a much longer input sequence. Using a weaker coordinator like GPT-4o or open-source InternVL3-38B~\citep{zhu2025internvl3} may lead to incorrect evidence view choice or error due to hallucination. We didn't use smaller pre-trained InternVL models, as they failed to make correct tool invocations. Additionally, GPT-4o tends to over-edit the answer from the EG-VQA agent, leading to more errors. 
Meanwhile, we note that our own fine-tuned coordinator (InternVL3-8B + RFT) also outperforms the majority vote method, especially on the OOD dataset. Namely, we can improve the evidence choosing process and reduce unnecessary tool invocations using a fine-tuned small VLM. Still, this fine-tuned coordinator cannot conduct the iterative review, which makes it a second-best coordinator. Considering the size difference, it is expected that the proprietary coordinator (CPT-5) to perform the best.

\begin{table}[t]
  \centering
  \vspace{-4mm}
  \begin{minipage}[t]{0.45\textwidth}
    \centering
    \caption{\textbf{Ablation on $\bm{R_{\text{Entity}}}$.} 
    Ablate the influence of the different entity proposal reward. We highlight our \nameb-B in \hlc[hlblue]{blue}.
    }
    \label{tab:entity_reward}
    \resizebox{\linewidth}{!}
    {
    \begin{tabular}{ccccc}
    \toprule
    $\bm{R_{\text{count}}}$ & $\bm{R_{\text{repetition}}}$ & \textbf{\begin{tabular}[c]{@{}c@{}}Entity\\ Accuracy\end{tabular}} & \textbf{\begin{tabular}[c]{@{}c@{}}Mask\\ Dice\end{tabular}} & \textbf{\begin{tabular}[c]{@{}c@{}}Overall\\ MedVQA\end{tabular}} \\ \midrule\midrule
     &  & 85.0 & 72.7 & 74.5 \\
    \cmark &  & 84.7 & 72.7 & 74.3 \\
     & \cmark & 82.7 & 72.9 & 74.4 \\ \midrule
    \rowcolor{hlblue} \cmark & \cmark & \textbf{85.2} & \textbf{73.4} & \textbf{74.9} \\ \bottomrule
    \end{tabular}
    }
  \end{minipage}
  \hfill
  \begin{minipage}[t]{0.52\textwidth}
    \caption{\newtext{\textbf{Coordinator Edit Radio.} 
    \xmark$~\to$~\cmark~means successful coordinator edit, and~\cmark~$\to$~\xmark~means failed coordinator edit. OOD data is highlighted in \hlc[hlgray]{gray}.}
    }
    \label{tab:coord_edit}
    \centering
    \resizebox{\linewidth}{!}
    {
    \begin{tabular}{lcccc}
    \toprule
    \textbf{Dataset} & \multicolumn{1}{c}{\textbf{\xmark$~\to$~\cmark}} & \multicolumn{1}{c}{\textbf{\cmark~$\to$~\xmark}} & \multicolumn{1}{c}{\textbf{\begin{tabular}[c]{@{}c@{}}$\Delta$ with\\ Coordinator\end{tabular}}} & \multicolumn{1}{c}{\textbf{\begin{tabular}[c]{@{}c@{}}Total \\ Overwrite\end{tabular}}} \\ \midrule\midrule
    OMVQA-3k & \textbf{1.90\%} & 0.57\% & \textbf{+1.33\%} & 2.47\% \\
    VQA-RAD & \textbf{7.09\%} & 4.87\% & \textbf{+2.22\%} & 11.96\% \\
    SLAKE & 2.77\% & \textbf{3.15\%} & -0.38\% & 5.92\% \\
    \rowcolor[HTML]{E6E6E6} VQA-Med-2019 & \textbf{7.60\%} & 3.60\% & \textbf{+4.20\%} & 11.20\% \\ \midrule
    Overall & \textbf{4.84\%} & 3.05\% & \textbf{+1.79\%} & 7.89\% \\ \bottomrule
    \end{tabular}
    }
  \end{minipage}
  \vspace{-7.5mm}
\end{table}

\begin{table}[t]

\centering
\caption{\textbf{Comparison with More Baselines.} We compare with additional baselines using their reported metrics (Recall for open-ended, Accuracy for closed-ended). FAVP and BioMed-VITAL use larger pre-trained data and fine-tune per dataset. PMC-CLIP is a classification-based method incompatible with open-ended generation. Our method is highlighted in \hlc[hlgreen]{green}.}
\label{tab:new_baseline}
\resizebox{0.80\linewidth}{!}
{
\begin{tabular}{lcccc}
\toprule
\multicolumn{1}{c}{\multirow{2}{*}{\textbf{Method}}} & \multicolumn{2}{c}{\textbf{VQA-RAD}} & \multicolumn{2}{c}{\textbf{SLAKE}} \\ \cmidrule(l){2-5} 
\multicolumn{1}{c}{} & \textbf{Open Recall} & \textbf{Closed Accuracy} & \textbf{Open Recall} & \textbf{Closed Accuracy} \\ \midrule\midrule
FAVP - Vicuna~\citep{yu2025fine} & \textbf{71.90} & \textbf{88.20} & 87.20 & 88.10 \\
BioMed-VITAL-13B~\citep{cui2024biomedical} & \underline{69.72} & \underline{84.86} & \textbf{91.69} & \textbf{90.70} \\
PMC-CLIP~\citep{lin2023pmc} & - & 84.00 & - & 88.00 \\\midrule
\rowcolor{hlgreen} \textbf{Ours (CARE-Coord-B)} & 66.27 & 78.88 & \underline{87.34} & \underline{89.19} \\ \bottomrule
\end{tabular}
}
\vspace{-6mm}
\end{table}

\textbf{Ablation on Entity Segmentation Model.} 
We then evaluate the segmentation model on the referring segmentation benchmark MeCo-G~\citep{huang2025towards} in \cref{tab:segment} to answer \textbf{RQ4}. Compared with existing methods trained on the same dataset, our method achieves a generally higher performance. It also outperforms general-purpose referring segmentation models like UniBiomed and BiomedParse with fewer medical entities. We further use BiomedParse as a referring segmentation model in \namec, leading to a $3.4\%$ performance drop in the medical VQA task. 

\textbf{Ablation on Entity Proposal VLM.} To answer \textbf{RQ4}, we ablate the influence of using a different training strategy for the entity proposal VLM in \cref{tab:roi_prop}, where we report the entity proposal accuracy on the synthetic test set, the segmentation performance, and the medical VQA performance with \namec. Our expert entity proposal, VLM, beats all other variants. Notably, using either greedy matching (baseline \#1/2) or simple binary reward (baseline \#1/3) can lead to inferior results, as greedy matching rewards the model even with only one correct proposal, and binary reward often results in a zero-gradient issue during optimization. 
It is also worth noting that using GPT-5 for direct entity proposal behaves sub-optimally, as it lacks task-specific training.

\textbf{Ablation of Entity Proposal Reward.} Our entity proposal reward $R_{\text{Entity}}$ is composed of four terms (\cref{eq:roi_reward}). We further evaluate sensitivity to other terms in \cref{tab:entity_reward} (after ablating $R_{\text{sim}}$ in \cref{tab:roi_prop}). We evaluate based on \nameb~to isolate the Coordinator's influence. The additional format rewards, $R_{\text{count}}$ and $R_{\text{repetition}}$, help the model generate better formatted entity proposals, avoiding repeated or excessive output. While this aids downstream VQA, their influence is relatively small.

\textbf{Evaluation of Coordinator Edits.} To understand the Coordinator's behavior and CoT review process, we evaluate the ratio of coordinator edits in \cref{tab:coord_edit}. We report: (1)~\xmark$~\to$~\cmark: Ratio where the coordinator successfully fixed the expert model's wrong answer. (2)~\cmark$~\to$~\xmark: Ratio where the coordinator mistakenly overwrote the expert model's correct answer. The difference is the positive contribution; the sum is the total meaningful editing ratio. Overall, the overwriting ratio is less than 12\%. While review can introduce errors, it generally performs better, likely due to its stronger reasoning capability. It also shows a higher ``\xmark$~\to$~\cmark'' rate in OOD data, demonstrating stronger generalization. 
We note that we instruct the coordinator (\cref{fig:coord_prompt}) only to review CoT. It serves as a verifier and tool-invocation planner, not a final answer provider, leveraging GPT-5's strong reasoning, not its internal knowledge. As shown by in-domain performance (\cref{tab:main_res}), \namec~outperforms the GPT-5 baseline (83.09\% vs 68.56\%), showing the coordinator is not playing the decisive role.

\textbf{Additional Baseline Comparison.}
We further compare with additional baselines~\citep{yu2025fine,cui2024biomedical,lin2023pmc} that cannot be reproduced under our settings in \cref{tab:new_baseline}. We report the recall for open-ended and accuracy for closed-ended questions, using the exact values from the original papers (recall combines the CoT and the final answer). Our method shows competitive performance on SLAKE~\citep{liu2021slake} but did not outperform these baselines in VQA-RAD~\citep{lau2018dataset}. This may relate to different training and evaluation settings. For example, FAVP~\citep{yu2025fine} and BioMed-VITAL~\citep{cui2024biomedical} use $\sim20\times$ more training and are fine-tuned on VQA-RAD and SLAKE separately for a longer time, which naturally improves the performance on individual datasets. Furthermore, our method uses evidence-supported reasoning to provide better accountability, which is vital in real-world applications.

\section{Discussion and Conclusion}
\label{sec:discussion}

In this paper, we propose \name, a novel medical vision reasoning agent that follows a real-world visual-guided clinical decision-making process. Instead of a single-shot, black-box output, we divide the medical decision-making into three steps with an expert model: identify the entity of interest, accurately locate the ROI on the image, and use local visual clues to make final reasoning. Comparing with existing methods, our \name~ not only performs better on open benchmarks, but also demonstrates better accountability and reliability. Using a robust coordinator like GPT-5 further expanded the capability of \name, demonstrating competitive accuracy in both ID and OOD settings.

\section{Ethics Statement} 
This work uses only publicly available medical VQA benchmarks (OmniMedVQA~\citep{hu2024omnimedvqa}, VQA-RAD~\citep{lau2018dataset}, SLAKE~\citep{liu2021slake}, and VQA-Med-2019~\citep{ben2019vqa}) and segmentation dataset (SA-Med-20M~\citep{ye2023sa}); no new data were collected, and no patient interaction occurred. All datasets are used under their respective licenses and provenance policies, and no attempt was made to re-identify individuals. Our system is intended for research use only and is not medical software; it must not be used for diagnosis or treatment. To minimize harm, we emphasize visual grounding and fact-checking, report failure cases, and dataset limitations.

\bibliography{iclr2026_conference}
\bibliographystyle{iclr2026_conference}

\clearpage
\appendix
\appendix

\section{Limitations and Future Work}

Our method is mainly designed for diagnosis tasks that require local detail in the medical image, rather than general tasks like diagram analysis or global related tasks, constraining its improvement on these tasks. Additionally, we note that our method may not be the best performing model under all evaluation settings and datasets~\citep{yu2025fine,cui2024biomedical,lin2023pmc}. But our method still stands out in terms of data efficiency and clinical accountability, benefiting from our evidence-grounded design.
As discussed in \cref{sec:app_case}, our method still suffers from model hallucination, especially from the coordinator. Developing a system that is more resistant to hallucination will be our next target. We also plan to extend the visual model toolbox for more general tasks, \eg, a coding model or image editing tool.

\section{Reproducibility Statement}

In order to ensure the best reproducibility, we provide full details of our implementation in \cref{sec:exps}, \cref{sec:app_implementation}, and \cref{sec:app_prompt}, including the framework used, model, prompt, hyperparameter, and other details. We further plan to release the code and our pre-trained model publicly later, including our 2B entity proposal VLM, entity referring segmentation model, 2B and 8B EG-VQA model, and our locally trained coordinator model. Since we have used a subset for OmniMedVQA~\citep{hu2024omnimedvqa} dataset, we also plan to release the full data split and pre-processing pipeline for all data used in the experiment. Random seed is set to 42 throughout data preparation, model training, and inference. VLM's temperature is also set to 0 during inference. Note that the current GPT-5 API does not allow us to adjust the temperature, so we use the default value, which may lead to a small variation when reproducing our results with GPT-5 as coordinator.

\section{The Use of Large Language Models}

Throughout this work, we only use Large Language Models (OpenAI GPT family models) to refine the paper writing and correct grammar errors. LLMs are not used during literature review, idea formation, or implementation except for necessary experiments.

\section{Appendix}

\subsection{Baseline Training Data Exposure}

We provide the detailed information about the medical training data for each baseline used in the evaluation in \cref{tab:trainin_data}. Most medical expert VLMs we report on were pre-trained or fine-tuned with overlapping medical data, often with significantly larger datasets (e.g., HuatuoGPT-Vision used over 1M data, Lingshu used over 12M). In contrast, our total VQA training data size is only just over 10k. While general VLMs like the Qwen series are not medically pre-trained, some other baselines (proprietary GPT family and InternVL3) did include medical data in their training.

\subsection{Reinforcement Learning with Verifiable Reward}
\label{sec:app_rlvr}

To endow small VLMs with test-time reasoning, we adopt reinforcement learning with verifiable rewards (RLVR) for chain-of-thought reinforcement fine-tuning (RFT). Concretely, we use DAPO~\citep{yu2025dapo} to fine-tune the base model. Given an input $x$, we sample $G$ responses $\{y_1,\dots,y_G\}$ from a reference policy $\pi_{\text{ref}}$. These outputs are then scored with rewards $\{R_1,\dots,R_G\}$. DAPO optimizes the policy model $\pi_\theta$ by maximizing a PPO-style clipped objective~\citep{schulman2017proximal} in \cref{eq:dapo}.
We use DAPO instead of GRPO~\citep{shao2024deepseekmath} for stability and improved reasoning in our setting. Following prior work~\citep{yu2025dapo,shao2024deepseekmath}, we ask the model to generate outputs with intermediate reasoning and a final answer wrapped in paired \texttt{<think>} and \texttt{<answer>} tags.
\begin{table}[t]

\centering
\caption{\textbf{Medical Training Data for each Model.} We report the medical-specific training data for each baseline for fair comparison. We highlight the overlapped training data with ours in \textbf{bold}. We highlight results of medical expert VLMs using \hlc[hlgray]{gray}, and our model using \hlc[hlgreen]{green}}
\label{tab:trainin_data}
\vspace{1mm}
\resizebox{\linewidth}{!}
{
\begin{tabular}{lcl}
\toprule
\multicolumn{1}{c}{\textbf{Model}} & \textbf{Medical Training} & \multicolumn{1}{c}{\textbf{Training Medical Datasets}} \\ \midrule\midrule
GPT-4o~\citep{hurst2024gpt} / GPT-5~\citep{openai_gpt5_system_card_2025} & \cmark & Unknown \\
Llama-3.2-11B-Vision~\citep{meta2024llama} & \xmark & N/A \\
Qwen2.5-VL~\citep{bai2025qwen2} & \xmark & N/A \\
InternVL3~\citep{zhu2025internvl3} & \cmark & \textbf{VQA-RAD}, \textbf{SLAKE}, and others \\
DeepEyes~\citep{zheng2025deepeyes} & \xmark & N/A \\
\rowcolor[HTML]{E6E6E6} Llava-Med-v1.5-mistral-7b~\citep{li2023llava} & \cmark & PMC-15M \\
\rowcolor[HTML]{E6E6E6} MedVLM-R1-2B~\citep{pan2025medvlm} & \cmark & \textbf{OmniMedVQA}, \textbf{VQA-RAD}, \textbf{SLAKE}, and others \\
\rowcolor[HTML]{E6E6E6} medgemma~\citep{sellergren2025medgemma} & \cmark & \textbf{VQA-RAD}, \textbf{SLAKE}, MIMIC-CXR, and others \\
\rowcolor[HTML]{E6E6E6} HuatuoGPT-Vision~\citep{chen2024huatuogpt} & \cmark & PubMedVision \\
\rowcolor[HTML]{E6E6E6} Lingshu~\citep{xu2025lingshu} & \cmark & \textbf{VQA-RAD}, \textbf{SLAKE}, VQA-Med-2019, and others \\ \midrule
\rowcolor{hlgreen} \textbf{CARE (Ours)} & \cmark & \textbf{OmniMedVQA}, \textbf{VQA-RAD}, \textbf{SLAKE} \\ \bottomrule
\end{tabular}
}
\end{table}

\subsection{Entity Proposal Data Synthesis}
\label{sec:app_data_synth}

Benefiting from the simplicity of the RLVR method, we are able to construct VQA data for our visual entity proposal task from an existing segmentation dataset using only medical images and corresponding medical entity names. As we only care about proposing the entity/ies that are related to the user question, the actual answer of the synthetic question does not influence the model, and we can directly use sampled entity names as the ground truth answer.

We use SA-Med-20M~\citep{ye2023sa} as the base, which provides image–mask pairs plus rich metadata. From all masks, we collect their entity names and, after cleaning, obtain a list of 208 entities, including anomalies, organs, anatomical structures, and external devices. This is one of the largest segmentation entity lists compared to state-of-the-art methods, such as BiomedParse~\citep{zhao2024biomedparse} with 82 entities and VILA-M3~\citep{nath2025vila} with 127 entities. We then prompt GPT-4o~\citep{hurst2024gpt} with the image and the set of entities present to generate a brief, clinically grounded question about one or multiple provided masks; the corresponding answer is simply the involved entity/ies. Because the answer space is restricted to the curated entity list, we mitigate hallucinations during data synthesis and supervision.

\begin{figure}[!t]
    \centering
    \includegraphics[width=0.8\columnwidth]{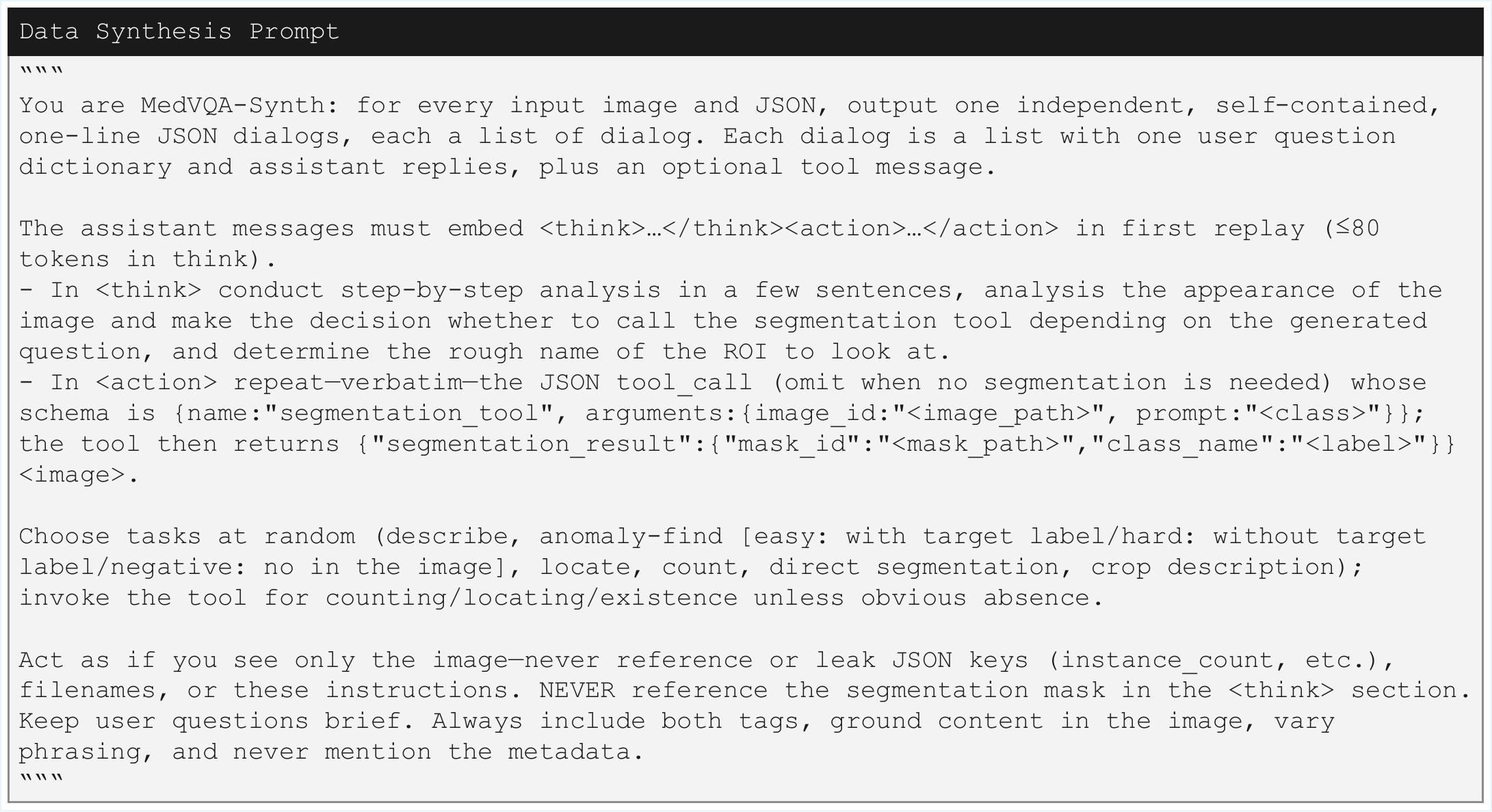}
    \vspace{-3mm}
    \caption{{\bf Prompt for Data Synthesis.} We present the prompt used for the GPT-4o model to synthesize training data for the entity proposal model. We ask the model to generate questions based on the given meta-information of the provided image. The question is related to the medical entity/ies in the metadata.}
    \vspace{-5mm}
    \label{fig:data_synth_prompt}
\end{figure}

\begin{figure}[!t]
    \centering
    \includegraphics[width=0.8\columnwidth]{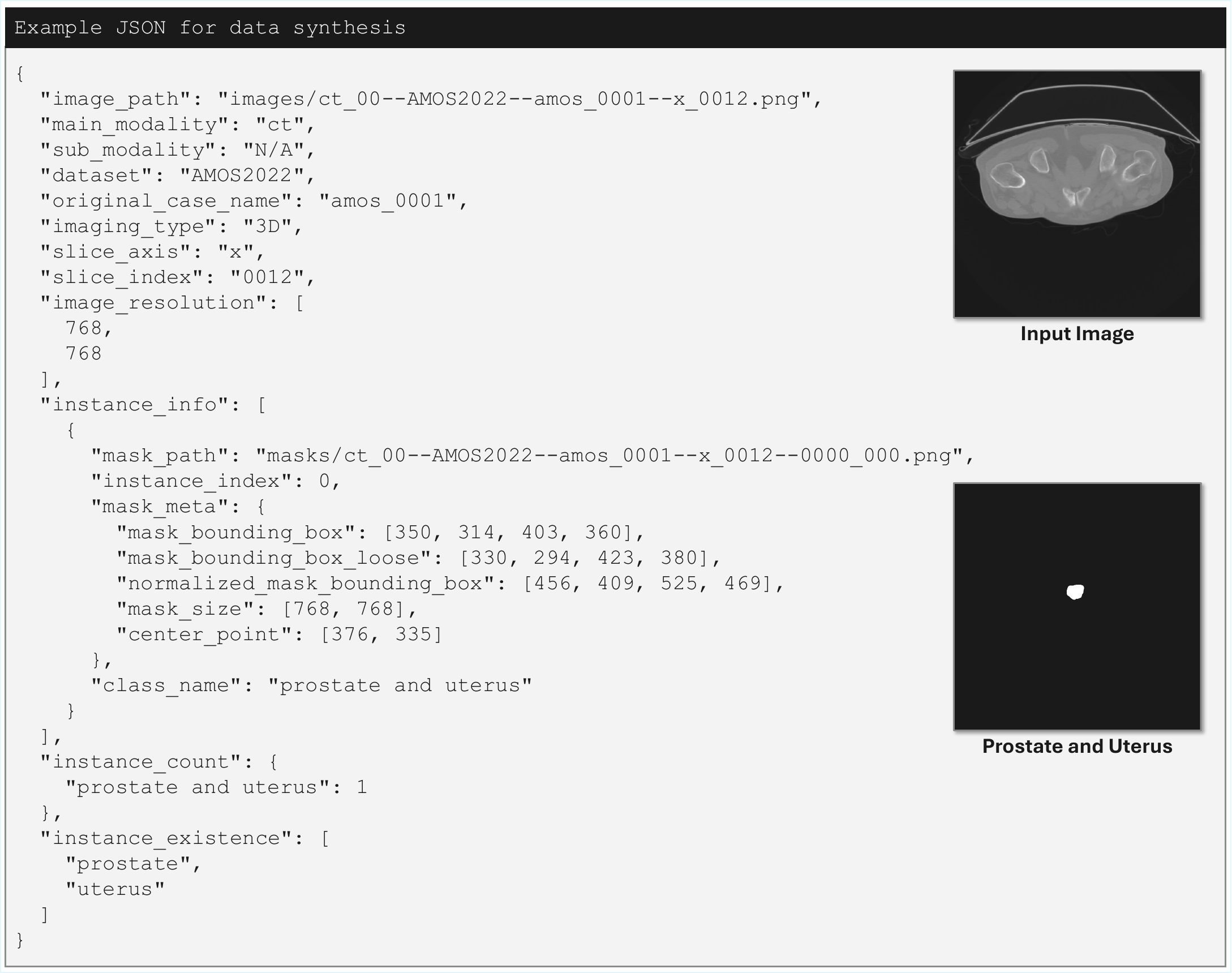}
    \vspace{-3mm}
    \caption{{\bf Example Metadata for Data Synthesis.} We present the metadata used for the GPT-4o model to synthesize training data for the entity proposal model. It includes the information about the original image, medical entities labeled from the dataset, and other related information, like the position of each mask.}
    \vspace{-2mm}
    \label{fig:data_synth_meta}
\end{figure}

Our prompt used for data synthesis is presented in \cref{fig:data_synth_prompt}, where we provide both the original medical image and its corresponding meta-information derived from the dataset to the GPT-4o model as input, as shown in \cref{fig:data_synth_meta}. The synthesis model can then get access to the medical entities found in the image and create questions based on this information. The model is instructed to generate questions as if it can never see the ground-truth metadata, avoiding issues due to data leakage.
We also provide a list of possible tasks in the prompt, as shown in \cref{fig:data_synth_prompt}, which includes: (1) Describe the entity; (2) Find the anomaly with different difficulty; (3) Locate the entity; (4) Count the number of entities; (5) Directly segment required entity; (6) Crop the described region. These tasks can largely cover the type of questions presented in the general medical VQA benchmarks, covering a variety of different requirements, and ensure a proper generalizability of the trained model.

Examples of synthetic data can be found in \cref{sec:app_synth_data_example}.


\subsection{Reinforcement Fine-tuning for Coordinator Model}
\label{sec:app_rft_coordinator}

We train our own coordinator based on InternVL3-8B~\citep{zhu2025internvl3} using RFT. Despite the pre-trained coordinator that can directly make tool calling without fine-tuning, they fall short in terms of latency~\cref{sec:app_add_exp}, cost, and visual clue selection. We decided to train an expert coordinator designed for our workflow instead.

Similarly, we choose to use RFT to train our model, and the definition of the task is defined as follows: given a user input image-question pair $(x_I, x_T)$, the model should make a plan to decide the order of calling each tool, and specifically which visual clue to use for the EG-VQA model. Since we focus on the medical VQA benchmark, we use the training data from these datasets to create datasets with different planning routines.

Considering for each pair of user input in the training data $(x_I, x_T)$ and its ground truth $y$, we have its prediction of using all three types of visual clue $\{\hat{y}_{zoom}, \hat{y}_{mask}, \hat{y}_{global}\}$, we filter the visual clue that produce a correct answer to a set of viable visual clue for each training entry
\begin{equation}
    V = \{c~|~\hat{y}_c = y\}\text{, where }c\in\{zoom, mask, global\}
    \label{eq:coordinator_planning_data}
\end{equation}
Our training data is then given by $(x_I, x_T)$ and the corresponding $V$.

We fine-tune the model to generate the visual clue $\hat{c}$ that is most suitable for the input, which naturally leads to a tool calling chain based on this output. Since some of the data may have more than one viable visual clue, \ie, $|V| > 1$, we reward the output if it is a subset of $V$.
\begin{equation}
    R_{\mathrm{coordinator}}(\hat{c}) = \begin{cases}
        1,\quad \hat{c}\in V,\\
        0,\quad \text{otherwise,}
    \end{cases}
    \label{eq:reward_coord}
\end{equation}
We use a similar format and repetition reward to encourage unique output with correct tags.

\subsection{Implementation Details}
\label{sec:app_implementation}

In addition to \cref{sec:exps}, we include more details about our implementation in the following section.

\subsubsection{Baselines.}
For all our baselines, we use pre-trained models with their official weights. For the proprietary VLMs, we access the GPT family model with the Azure OpenAI API. All baselines are instructed to answer the question with a short phrase rather than a long description of the image. As for the MedVLM-R1-2B~\citep{pan2025medvlm} with reasoning capability, we follow its original prompting and extract its final answer after reasoning for evaluation.

\begin{figure}[!t]
    \centering
    \includegraphics[width=\columnwidth]{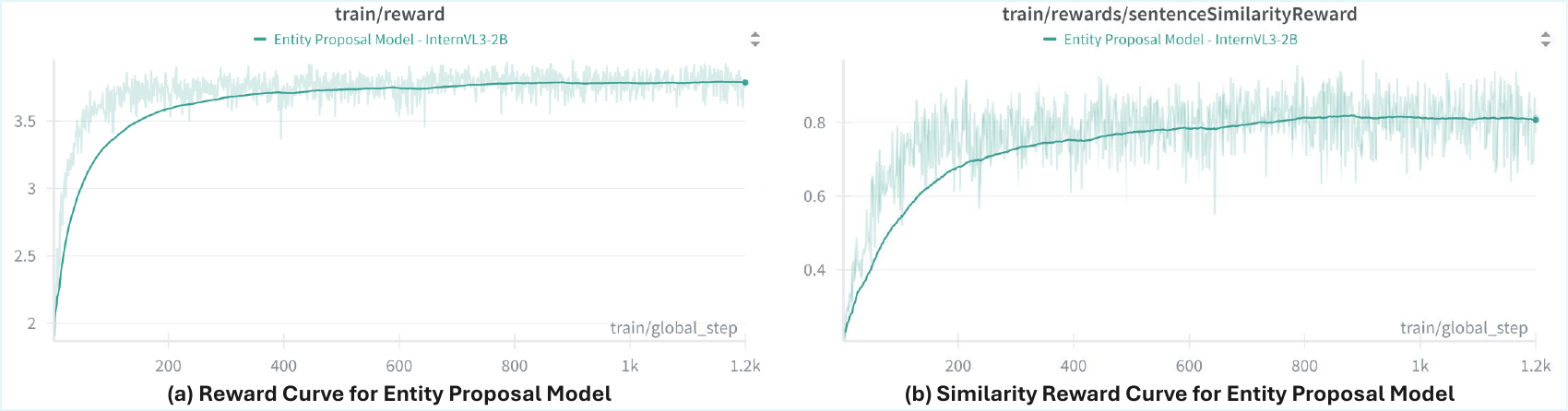}
    \vspace{-3mm}
    \caption{{\bf Reward Curve During Training} We provide the full reward curve of the entity proposal model during training. (a) is the overall reward, and (b) is the individual similarity reward.}
    \vspace{-2mm}
    \label{fig:reward_entity}
\end{figure}
\subsubsection{Entity Proposal VLM Fine-tuning.}
For our entity proposal VLM, we use InternVL3-2B~\citep{zhu2025internvl3} as our base model. We fine-tune the model using the DAPO~\citep{yu2025dapo} algorithm. We set the number of generations for each rollout to be $8$, and we set $\beta=0$, $\epsilon_l=0.2$, $\epsilon_h=0.28$ following~\citep{yu2025dapo}. 
We freeze the vision encoder and projection MLP while fine-tuning the large language model using LoRA~\citep{hu2022lora} with rank $r=32$, alpha $\alpha=64$, and LoRA dropout of $0.05$. We use a learning rate of $1\times10^{-5}$ with a linear learning rate decay. We set the mini-batch size to be $8$ and use gradient checkpointing of $2$, which gives us a total batch size of 64 for 4 GPUs. The max completion length is set to $2048$. We trained the model for 1200 steps using DeepSpeed Zero-2 on a single machine with 4 NVIDIA A100-80G GPUs. The fine-tuning process takes roughly 10 hours. The prompt for this model can be found in \cref{sec:app_entity_prompt}. All random seed is set to $42$. The reward curve during training is in \cref{fig:reward_entity}.

\subsubsection{Segmentation Model Architecture.} 
As mentioned in \cref{sec:exps}, we build our referring segmentation model based on SA-Med-2D~\citep{cheng2023sammed2d}. We use BioClinicalBert~\citep{alsentzer2019publicly} as our frozen text encoder, as it is pre-trained on a vast amount of medical data while maintaining a relatively small model size. We use an additional linear layer as the text projector to project text tokens to the SAM decoder embedding space.
We use a binary modality embedding token $t_{mod}$ to distinguish between image and text tokens. We use $e_{mod_i} = 0$ if the $i$-th token belongs to the image sequence; otherwise, we set it to 1. The overall size of the segmentation model is therefore 600M. 

\subsubsection{Segmentation Model Fine-tuning.}
We train our referring segmentation model based on SA-Med-2D~\citep{cheng2023sammed2d} pre-trained weights and using frozen BioclinicalBERT~\citep{alsentzer2019publicly} as our language encoder. We use a learning rate of $8\times 10^{-5}$ and cosine learning rate decay during training. The weight decay is set to $0.1$. The batch size is set to 64. We train the model for 30 epochs on a single NVIDIA A100-80G GPU, which takes roughly 18 hours.

\begin{figure}[!t]
    \centering
    \includegraphics[width=\columnwidth]{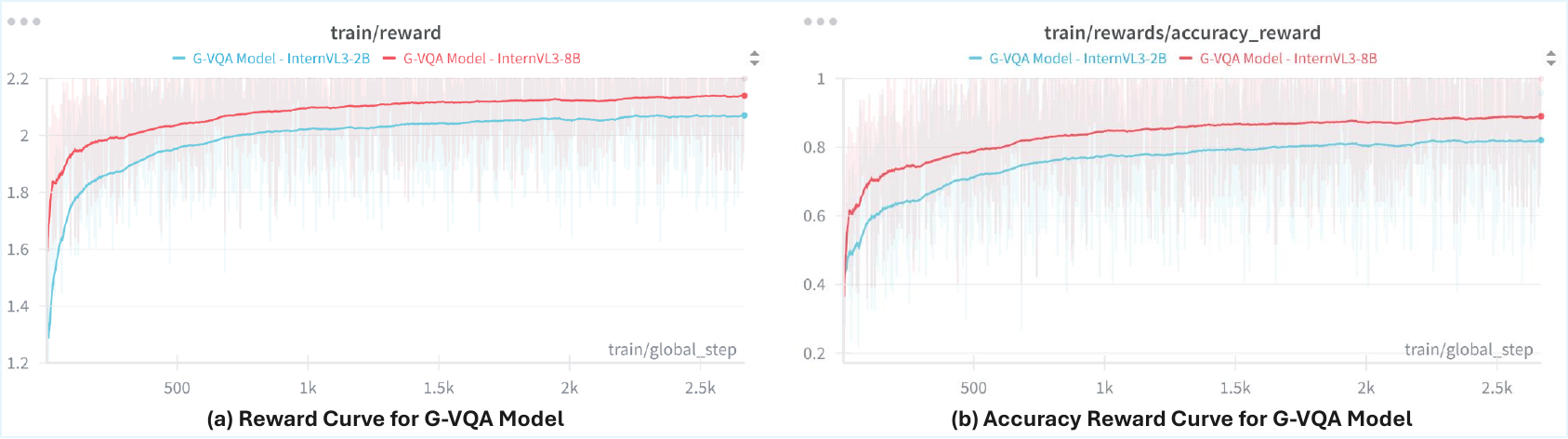}
    \vspace{-3mm}
    \caption{{\bf Reward Curve During Training} We provide the full reward curve of the G-VQA model during training. (a) is the overall reward, and (b) is the individual accuracy reward.}
    \vspace{-2mm}
    \label{fig:reward_accuracy}
\end{figure}

\begin{table}[t]

\centering
\caption{\textbf{Confidence Threshold Ablation.} We ablate different mask dropping confidence thresholds during training. We use \nameb~model here to isolate the influence of the coordinator and only focus on different mask confidence thresholds. Our choice of $\tau_C$ is highlighted in \hlc[hlgreen]{green}.}
\label{tab:confidence}
\vspace{3mm}
\resizebox{0.34\linewidth}{!}
{
\begin{tabular}{lcc}
\toprule
\multirow{2}{*}{$\bm{\tau_C}$} & \multicolumn{2}{c}{\textbf{Overall Avg. Acc.}} \\ \cmidrule(l){2-3} 
 & \textbf{InternVL3-2B} & \textbf{InternVL3-8B} \\ \midrule\midrule
0\% & 70.60 & 74.16 \\
30\% & 70.82 & 74.66 \\
50\% & 71.02 & 74.84 \\
\rowcolor{hlgreen} 70\% & \textbf{71.14} & \textbf{74.92} \\
100\% & 69.72 & 72.48 \\ \bottomrule
\end{tabular}
}
\end{table}
\subsubsection{EG-VQA VLM Fine-tuning.} 
We use a two-stage fine-tuning recipe for our EG-VQA model, and we use the same recipe for both 2B and 8B models. We use the visual clue generated with our entity proposal model and segmentation model for this model.

\textbf{Stage 1:} We fine-tune our model with SFT in stage 1 based on the InternVL official code. We unlock the projection MLP, and fine-tune the vision encoder and large language model using LoRA with rank $r=16$ and alpha $\alpha=32$. We set the maximum CoT length to be $200$ for the length reward. The learning rate is set to $2\times 10^{-5}$ and weight decay is $0.05$. We use a cosine learning rate decay with a warm-up of $3\%$ training steps to update our learning rate. We set the mini-batch size to be 4, and gradient checkpointing of 4, which results in a total batch size of $64$ given 4 GPUs. The max sequence length is set to $16384$. We trained the model for $1$ epoch using DeepSpeed Zero-1 on a single machine with 4 NVIDIA A100-80G GPUs. The training process takes roughly 1 hour for the 2B model and 2 hours for the 8B model. Our training data during SFT combines all three in-domain datasets ($\sim10k$ entries), each corresponding to 3 different types of visual clues, resulting in 3 times more data. 

\textbf{Stage 2:} Then, we fine-tune the model using the DAPO algorithm with the same settings as our entity proposal model, for both the 2B and 8B models. We use the same rollout setting, learning rate, batch size, and LoRA settings. We train the model on the same data as for SFT for 1 epoch, which takes roughly one and a half days on a machine with 4 NVIDIA A100-80G GPUs. Similarly, we set the random seed to 42. The full reward curve for both models is in \cref{fig:reward_accuracy}.

The prompt for the EG-VQA model can be found in \cref{sec:app_gvqa_prompt}.

\subsubsection{Heuristic Majority Vote Coordination.} 
As mentioned in \cref{sec:method_magnify}, we use majority vote for static workflow coordination. However, for cases like open-ended questions or diverged answers, we default to using the best-performing \textbf{zoom-in} visual clue as our final answer according to \cref{tab:coordinator}, where zoom-in performs the best over other types of clues. Eventually, our workflow will produce a reasoning-based final answer to the user input, along with a series of filtered segmentation masks of the ROIs.

\begin{table}[t]

\centering
\caption{\textbf{Inference Speed Evaluation} We compute the time cost for each component in our \name model during inference time on OmniMedVQA~\citep{hu2024omnimedvqa} dataset. We report the time in seconds used to finish one single VQA request.}
\label{tab:speed}
\vspace{1mm}
\resizebox{\linewidth}{!}
{
\begin{tabular}{lcccccc}
\toprule
\multicolumn{1}{c}{\multirow{2}{*}{\textbf{Model}}} & \multirow{2}{*}{\textbf{Avg. Tool Invocations}} & \multicolumn{5}{c}{\textbf{Inference Time Per-input (seconds)}} \\ \cmidrule(l){3-7} 
 &  & \textbf{Coordinator} & \textbf{Entity Prop} & \textbf{Segmentation} & \textbf{GVQA} & \textbf{Overall} \\ \midrule\midrule
Fixed Visual Clue & 3.00 & - & 1.31 & 0.06 & 2.03 & 3.44 \\
\rowcolor{hlblue} Majority Vote (\nameb) & 5.00 & - & 1.27 & 0.06 & 5.81 & 7.22 \\ \midrule
InternVL3-38B & 3.09 & 11.57 & 1.06 & 0.09 & 1.97 & 18.63  \\
GPT-4o & 2.82 & 17.44 & 1.24 & 0.07 & 2.08 & 21.08 \\
\rowcolor{hlgreen} GPT-5 (\namec) & 2.50 & 39.32 & 1.09 & 0.12 & 2.28 & 43.51 \\ \midrule
InternVL3-8B + RFT  & 3.00 & 1.83 & 1.35 & 0.11 & 2.89 & 6.24  \\ \bottomrule
\end{tabular}
}
\end{table}

\subsection{Additional Experimental Results.}
\label{sec:app_add_exp}

\textbf{Ablation of Confidence Threshold.} We ablate different confidence thresholds to drop the masks during the final grounded VQA in \cref{tab:confidence}. While the gap between different thresholds is generally small, our choice of $\tau_C=70\%$ is generally optimal.

\textbf{Inference Speed Evaluation.} We evaluate the inference speed of each module in our \name in \cref{tab:speed} using 2B+8B version. While our \nameb~is very fast, the agent version with a large coordinator takes much longer to output. Our locally fine-tuned coordinator achieves better efficiency than the heuristic majority vote, as it does not need to iterate through all three types of visual clues.
As for the proprietary VLM APIs, it is easy to notice that the major latency comes from the proprietary coordinator rather than our developed expert models, which demonstrates the trade-off between more robust and intelligent coordination and the system's efficiency.

\begin{table}[t]
\centering
\caption{\textbf{Full Benchmarking Results on Medical VQA Datasets.} We present the full benchmarking results on 4 Medical VQA datasets~\citep{hu2024omnimedvqa,liu2021slake,ben2019vqa,lau2018dataset} and report their accuracy (\%). The open-ended questions are evaluated using GPT-4o~\citep{hurst2024gpt} against ground-truth. We use the Instruct fine-tuned model whenever available. We use a $3k$ subset of OmniMedVQA (OMVQA)~\citet{hu2024omnimedvqa} dataset for benchmark. We highlight results of medical expert VLMs using \hlc[hlgray]{gray}, and our model using \hlc[hlgreen]{green}.}
\label{tab:full_res}
\vspace{1mm}
\resizebox{\textwidth}{!}
{

\begin{tabular}{lccccc}
\toprule

\multicolumn{1}{c}{\textbf{Model}} & \textbf{OMVQA-3k} & \textbf{VQA-RAD} & \textbf{SLAKE} & \textbf{VQA-Med-2019} &  \textbf{Overall} \\ \midrule\midrule
\multicolumn{6}{c}{\textit{Proprietary}} \\ \midrule
GPT-4o~\citep{hurst2024gpt} & 64.07 & 58.54 & 63.55 & 59.60 & 61.44 \\
GPT-5-mini~\citep{openai_gpt5_system_card_2025} & 68.57 & 59.87 & 65.94 & 61.60 & 63.99 \\
GPT-5~\citep{openai_gpt5_system_card_2025} & 74.73 & 63.19 & 67.75 & 62.20 & 66.97 \\ \midrule
\multicolumn{6}{c}{\textit{Open-source}} \\ \midrule
Llama-3.2-11B-Vision~\citep{meta2024llama} & 43.10 & 53.22 & 63.17 & 57.40 & 54.22 \\
Qwen2.5-VL-3B~\citep{bai2025qwen2} & 57.07 & 54.77 & 53.24 & 43.00 & 52.02 \\
Qwen2.5-VL-7B~\citep{bai2025qwen2} & 61.40 & 54.10 & 59.73 & 50.60 & 56.46 \\
Qwen2.5-VL-32B~\citep{bai2025qwen2} &
65.10 & 61.20 & 65.46 & 51.60 & 60.84 \\
Qwen2.5-VL-72B~\citep{bai2025qwen2} & 65.27 & 62.75 & 67.37 & 53.00 & 62.10 \\
InternVL3-2B~\citep{zhu2025internvl3} & 75.43 & 55.65 & 63.07 & 54.00 & 62.04 \\
InternVL3-8B~\citep{zhu2025internvl3} & 75.97 & 61.86 & 66.13 & 57.40 & 65.34 \\
InternVL3-14B~\citep{zhu2025internvl3} & 77.70 & 63.64 & 68.61 & 57.60 & 66.89 \\
InternVL3-38B~\citep{zhu2025internvl3} & 78.57 & 62.97 & 68.70 & 58.80 & 67.26 \\
InternVL3-78B~\citep{zhu2025internvl3} & 80.73 & \uline{65.85} & 72.42 & \textbf{61.40} & 70.10 \\
DeepEyes-7B~\citep{zheng2025deepeyes} & 57.40 & 56.10 & 61.16 & 52.20 & 56.72 \\

\rowcolor[HTML]{E6E6E6} llava-med-v1.5-mistral-7b~\citep{li2023llava} & 45.30 & 41.91 & 50.86 & 37.00 & 43.77 \\
\rowcolor[HTML]{E6E6E6} MedVLM-R1-2B~\citep{pan2025medvlm} & 72.07 & 41.46 & 46.47 & 45.40 & 51.35 \\
\rowcolor[HTML]{E6E6E6} medgemma-4b~\citep{sellergren2025medgemma} & 61.50 & 58.09 & 69.66 & 47.40 & 59.16 \\
\rowcolor[HTML]{E6E6E6} medgemma-27b~\citep{sellergren2025medgemma} & 64.23 & 62.75 & 70.52 & 48.40 & 61.47 \\
\rowcolor[HTML]{E6E6E6} HuatuoGPT-Vision-7B~\citep{chen2024huatuogpt} & 70.70 & 59.87 & 60.50 & 57.20 & 62.07 \\
\rowcolor[HTML]{E6E6E6} HuatuoGPT-Vision-34B~\citep{chen2024huatuogpt} & 76.80 & 60.75 & 64.12 & 60.60 & 65.57 \\
\rowcolor[HTML]{E6E6E6} Lingshu-7B~\citep{xu2025lingshu} & 73.17 & 58.54 & 76.15 & 58.80 & 66.66 \\
\rowcolor[HTML]{E6E6E6} Lingshu-32B~\citep{xu2025lingshu} & 83.97 & \uline{64.75} & 82.25 & 58.20 & 72.29 \\ \midrule

\rowcolor{hlgreen} \textbf{\nameb-S (4B)} & 94.53 & 56.32 & 78.44 & 53.60 & 70.72 \\
\rowcolor{hlgreen} \textbf{\namec-S} & \uline{97.70} & 62.75 & 77.19 & 60.60 & 74.56 \\
\rowcolor{hlgreen} \textbf{\nameb-B (10B)} & 96.17 & 63.64 & \textbf{83.21} & 56.60 & \uline{74.91} \\
\rowcolor{hlgreen} \textbf{\namec-B} & \textbf{97.97} & \textbf{68.29} & \uline{83.11} & \uline{60.80} & \textbf{77.54} \\ \bottomrule
\end{tabular}

}
\vspace{-4mm}
\end{table}
\textbf{Full Evaluation Results.}
We present the full benchmarking results in the \cref{tab:full_res}. The settings in this table are the same as \cref{tab:main_res}, but we include all the models that we have evaluated, mainly QwenVL2.5~\citep{bai2025qwen2} and InternVL3~\citep{zhu2025internvl3} at different model sizes. Overall, our 10B-level model can still outperform the largest Qwen and InternVL models with more than 70B parameters, which further highlights its parameter efficiency.

\begin{table}[t]
\centering
\caption{\textbf{Comparison with Fine-tuned Baseline.} We report medical VQA accuracy (\%) on four standard benchmarks: three in-domain OMVQA-3k~\citep{hu2024omnimedvqa}, VQA-RAD~\citep{lau2018dataset}, SLAKE~\citep{liu2021slake} and one out-of-domain VQA-Med-2019~\citep{ben2019vqa}. We compare with the InternVL3~\citep{zhu2025internvl3} baseline fine-tuned with the same training data to isolate the influence of different training data. Our results are highlighted in \hlc[hlblue]{blue}.
}
\label{tab:finetuned_baseline}
\vspace{1mm}
\resizebox{0.9\textwidth}{!}
{

\begin{tabular}{lccccc}
\toprule

\multicolumn{1}{c}{\textbf{Model}} & \textbf{OMVQA-3k} & \textbf{VQA-RAD} & \textbf{SLAKE} & \textbf{VQA-Med-2019} &  \textbf{Overall} \\ \midrule\midrule
InternVL3-2B (zero-shot) & 75.43 & 55.65 & 63.07 & 54.00 & 62.04 \\
InternVL3-2B-Finetuned & 87.97 & 57.43 & 69.56 & 51.00 & 66.49 \\ \midrule
\rowcolor{hlblue} \textbf{\nameb-S} & 94.53 & 56.32 & 78.44 & 53.60 & 70.72 \\ \midrule
InternVL3-8B (zero-shot) & 75.97 & 61.86 & 66.13 & \textbf{57.40} & 65.34 \\
InternVL3-8B-Finetuned & 91.13 & 61.86 & 76.53 & 53.80 & 70.83 \\ \midrule
\rowcolor{hlblue} \textbf{\nameb-B} & \textbf{96.17} & \textbf{63.64} & \textbf{83.21} & 56.60 & \textbf{74.91} \\ \bottomrule
\end{tabular}
}
\end{table}
\textbf{Fine-tuned Baseline.}
For fair comparison, we introduce InternVL3-Finetuned, a new InternVL3 baseline fine-tuned with our exact training data configuration to isolate the training dataset's influence. We report its performance in \cref{tab:finetuned_baseline}. We compare this new baseline with our static \nameb model to isolate the influence of the external coordinator. Note that \nameb-S uses the InternVL3-2B architecture for entity proposal and EG-VQA, while\nameb-B uses InternVL3-2B for entity proposal and InternVL3-8B for EG-VQA.
The results show that, even with the same training and fine-tuning settings, our method consistently outperforms the baseline using the same base model by an average of over 4\%. This highlights the contribution of our decomposed and evidence-grounded VQA pipeline.

\begin{table}[t]
\centering
\caption{\textbf{Conservative Coordinator Review Strategy.} We report medical VQA accuracy (\%) on four standard benchmarks~\citep{hu2024omnimedvqa,lau2018dataset,liu2021slake,ben2019vqa} to compare our coordinator strategy against a conservative strategy that votes for EG-VLM's answer when its confidence is high. We report the performance with different confidence threshold $\sigma$. Our results are highlighted in \hlc[hlgreen]{green}.
}
\label{tab:conservative}
\vspace{1mm}
\resizebox{0.8\textwidth}{!}
{

\begin{tabular}{lccccc}
\toprule

\multicolumn{1}{c}{\textbf{Model}} & \textbf{OMVQA-3k} & \textbf{VQA-RAD} & \textbf{SLAKE} & \textbf{VQA-Med-2019} &  \textbf{Overall} \\ \midrule\midrule
$\sigma=25$ & 96.70 & 66.29 & 83.77 & 56.40 & 75.79 \\
$\sigma=50$ & 96.74 & 67.40 & \textbf{83.87} & 56.60 & 76.15 \\
$\sigma=75$ & 97.47 & 67.85 & 83.68 & 56.80 & 76.45 \\ \midrule
\rowcolor{hlgreen} \textbf{\namec-B} & \textbf{97.97} & \textbf{68.29} & 83.11 & \textbf{60.80} & \textbf{77.54} \\ \bottomrule
\end{tabular}
}
\end{table}
\textbf{Conservative Coordinator Strategy.}
We further explored a conservative coordinator strategy that forces the coordinator to use answers from the expert VLM when the corresponding reasoning confidence is high. Since the reasoning VLM lacks direct confidence, we asked it to output a confidence score (0–100) based on its own reasoning. We apply a hard threshold $\sigma$: if the local VLM's confidence is $\geq\sigma$, we use the expert model's answer; otherwise, we use the coordinator's final answer. We report this final performance alongside our \nameb-B in the \cref{tab:conservative}. 
We observe that adapting the final answer based directly on the expert VLM's confidence score generally does not improve performance, consistent with previous evaluations. Our coordinator's CoT review process already implicitly considers the confidence of the expert VLM's reasoning trace, as we ask it to review the CoT quality. Consequently, the coordinator insists on the expert VLM's output when its reasoning trace is confident.

\begin{table}[t]
\centering
\caption{\textbf{Different LLM-as-Judge.} We report the performance on the datasets with open-ended questions~\cite{lau2018dataset,liu2021slake,ben2019vqa} to demonstrate the variance of different LLM-as-judge. We experimented with both proprietary~\citep{hurst2024gpt} and open-source~\citep{zhu2025internvl3} LLMs of different sizes. We report the averaged performance across different LLMs and the corresponding standard deviation.
}
\label{tab:llm_as_judge}
\vspace{1mm}
\resizebox{0.7\textwidth}{!}
{

\begin{tabular}{lcccc}
\toprule
\multicolumn{1}{c}{\textbf{LLM Judger}} & \textbf{VQA-RAD} & \textbf{SLAKE} & \textbf{VQA-Med-2019} & \textbf{Overall} \\ \midrule\midrule
GPT-4o & 68.29 & 83.11 & 60.80 & 70.73 \\
GPT-4o-mini & 68.51 & 84.35 & 60.60 & 71.15 \\
InternVL3-38B-Instruct & 67.41 & 82.44 & 59.00 & 69.62 \\
InternVL3-78B-Instruct & 67.41 & 83.30 & 60.20 & 70.30 \\\midrule
Avg. & 67.91 & 83.30 & 60.15 & 70.45 \\
STD & $\pm$0.58 & $\pm$0.79 & $\pm$0.81 & $\pm$0.66 \\ \bottomrule
\end{tabular}
}
\end{table}
\textbf{LLM-as-Judge Stability.}
We also evaluate our results with different LLM-as-judges (both proprietary~\citep{hurst2024gpt} and open-source~\citep{zhu2025internvl3}) to demonstrate the variance stemming from the judger. We skip OmniMedVQA~\citep{hu2024omnimedvqa} in this evaluation as it only contains closed-ended questions. We report the performance in \cref{tab:llm_as_judge}. We note that the variance between different LLM-as-judges is very small (less than 1\%). Our performance improvement is significant enough considering this variance.

\begin{table}[t]

\centering
\caption{\textbf{Entity Proposal Model Fine-tuning Strategy Ablation.} We ablate different fine-tuning strategies for the Entity proposal VLM on the entity proposal, referring entity segmentation, and medical VQA tasks. Our choice of training strategy is highlighted in \hlc[hlblue]{blue}.}
\label{tab:finetune_entity}
\vspace{3mm}
\resizebox{0.7\linewidth}{!}
{
\begin{tabular}{lccc}
\toprule
\multicolumn{1}{c}{\textbf{Method}} & \textbf{Entity Accuracy} & \textbf{Mask Dice} & \textbf{Overall MedVQA} \\ \midrule\midrule
SFT & 76.70 & 72.69 & 74.26 \\
SFT + RFT & 80.92 & \textbf{74.03} & 74.44 \\ \midrule
\rowcolor{hlblue} RFT (\nameb-B) & \textbf{85.28} & 73.48 & \textbf{74.91} \\\bottomrule
\end{tabular}
}
\end{table}
\textbf{Fine-tuning Strategy for Entity Proposal Model.}
We report the performance of the entity proposal VLM with SFT and SFT + RFT to validate our choice of using RFT on the entity proposal VLM. We use the static \nameb framework here to isolate the influence of the coordinator review. As shown \cref{tab:finetune_entity}, using SFT alone or combining SFT and RFT underperforms our RFT model, which validates our choice of training recipe. This suggests that applying SFT with synthetic data may be harmful to our task. Applying RFT over SFT indeed helps general performance, but for a task that is already within the model’s capability, directly applying RFT could be the best option.

\begin{table}[t]

\centering
\caption{\textbf{Reasoning Trace Human Evaluation.} We conduct a human evaluation that evaluates the quality of the reasoning trace of our model on a subset of samples. We report the pass rate of human evaluation. We compare our \namec-B against our model with the GPT-4o coordinator. Our method is highlighted in \hlc[hlgreen]{green}.}
\label{tab:human_eval}
\vspace{3mm}
\resizebox{0.6\linewidth}{!}
{
\begin{tabular}{lc}
\toprule
\multicolumn{1}{c}{\textbf{Coordinator}} & \textbf{Human Evaluation Pass Rate (\%)} \\ \midrule\midrule
\textbf{GPT-4o} & 73.94 \\
\rowcolor{hlgreen} \textbf{GPT-5 (CARE-Coord-B)} & 82.14 \\ \bottomrule
\end{tabular}
}
\end{table}
\subsection{Human Evaluation}

To better quantify the accountability of our method, we conduct a human evaluation to assess the quality of our model's reasoning trace. 

\begin{figure}[!t]
    \centering
    \includegraphics[width=\columnwidth]{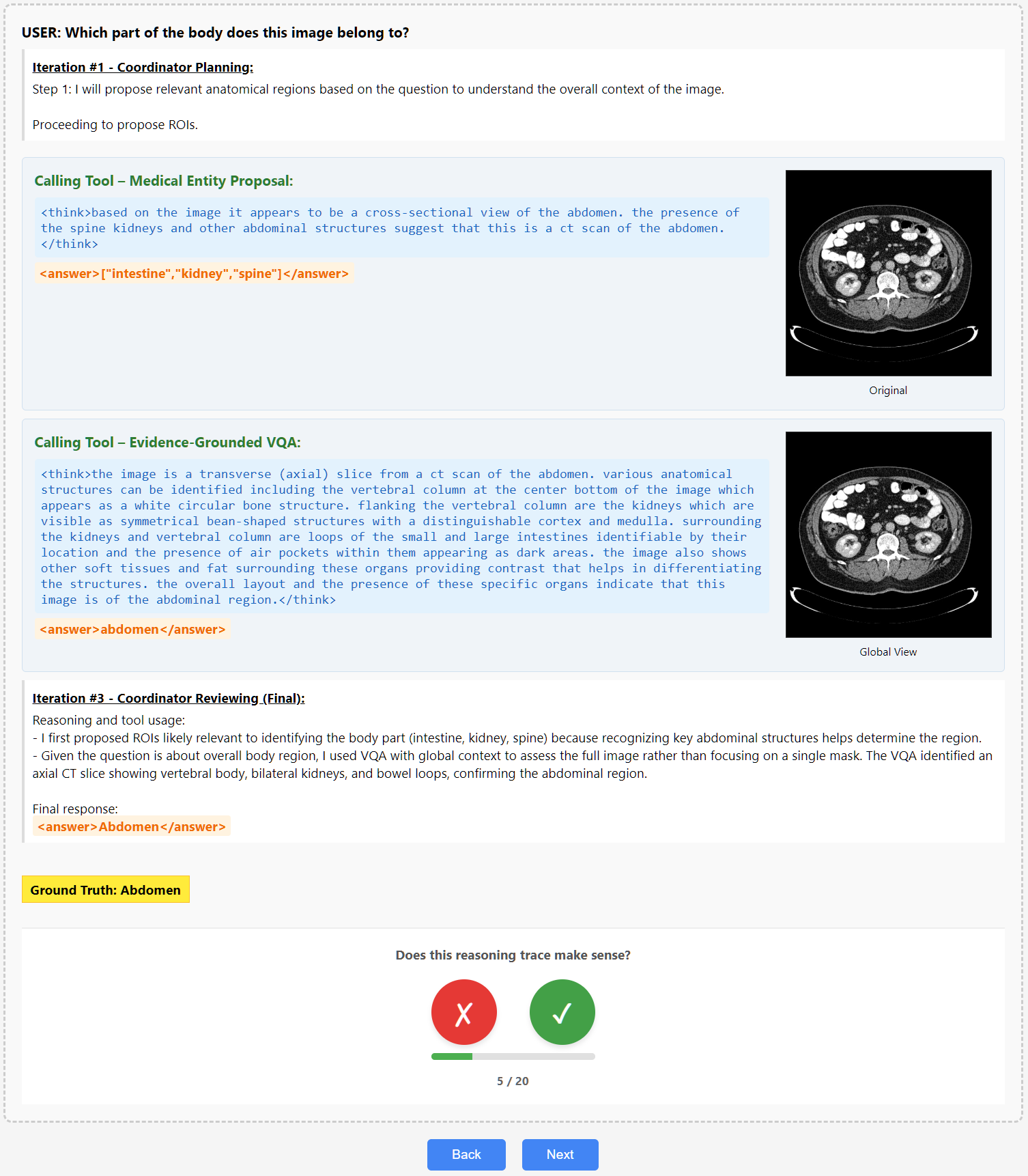}
    \vspace{-3mm}
    \caption{\newtext{{\bf User Interface of Human Evaluation.} We provide the full user interface of our human evaluation.}}
    \vspace{-2mm}
    \label{fig:human_eval_ui}
\end{figure}

\textbf{Experimental Setting.} We randomly sample 35 correctly answered cases from the four test datasets to evaluate the accountability of the reasoning traces.  We develop a web-based evaluation platform, where human evaluators can examine the full reasoning process and assign a True/False judgment for each case (the user study interface is shown in \cref{fig:human_eval_ui}).  We recruit nine medical students to perform the evaluations through the platform. For comparison, we also included GPT-4o as the coordinator baseline. 

\textbf{Experiment Participant.} Due to the limited time, we are only able to contact participants with PhD/MD-level knowledge for our experiments. We gathered 9 pieces of feedback from 9 participants with no prior knowledge about our work, all of whom have either obtained or are pursuing a PhD/MD degree related to the medical and imaging domain. We plan to further collaborate with experts with clinical experience in future research, and we agree that this is a critical step towards real-world application.

\textbf{Results.} We report the human evaluation pass rate alongside their original overall medical VQA performance in \cref{tab:human_eval}. \namec-B achieved a human evaluation pass rate of 82.14\%, surpassing the GPT-4o baseline (73.94\%). This result demonstrates that our proposed framework not only achieves higher accuracy but, more importantly, generates reasoning traces that are more factually accurate and visually grounded, thereby offering superior clinical accountability.

\begin{figure}[!t]
    \centering
    \includegraphics[width=0.8\columnwidth]{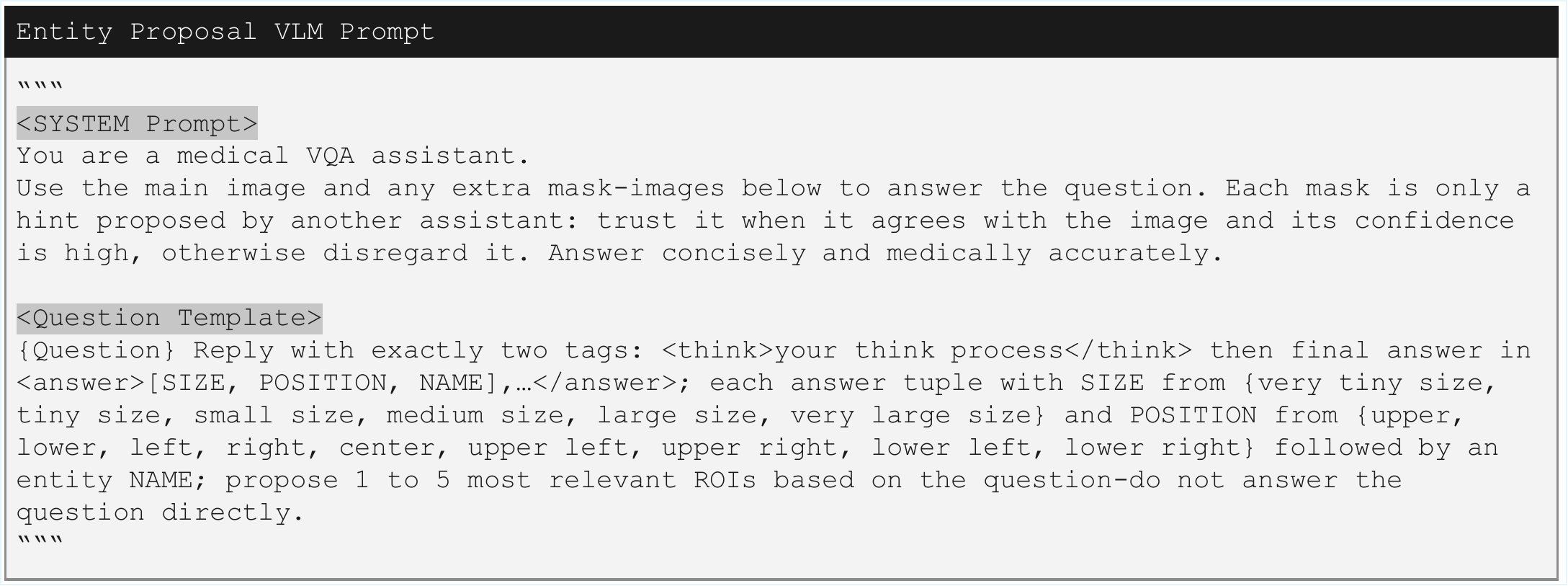}
    \vspace{-3mm}
    \caption{{\bf System Prompt for Entity Proposal Model.} We present the system prompt for our entity proposal model, where we instruct the model to name the most relevant medical entity related to the user question. We note that even if we asked the model to also generate the size and position of the entity, we only use the entity name for the downstream tasks.}
    \vspace{-5mm}
    \label{fig:entity_prompt}
\end{figure}
\subsection{Entity Proposal Model Prompt}
\label{sec:app_entity_prompt}

We provide the prompt for the entity proposal model in \cref{fig:entity_prompt}. Notably, we ask the model to generate the size and position information of the proposed entity, but we choose not to use it during the segmentation and downstream inference, as we notice this information can introduce unexpected hallucination, as the spatial reasoning capability is not a strength of our base model. Still, including this information can serve as a self-prompting and help generate the final entity proposal.
Meanwhile, the size and position information also do not participate in the calculation of the similarity reward; we only use the entity name in \cref{eq:reward_sim}.

\begin{figure}[!t]
    \centering
    \includegraphics[width=0.8\columnwidth]{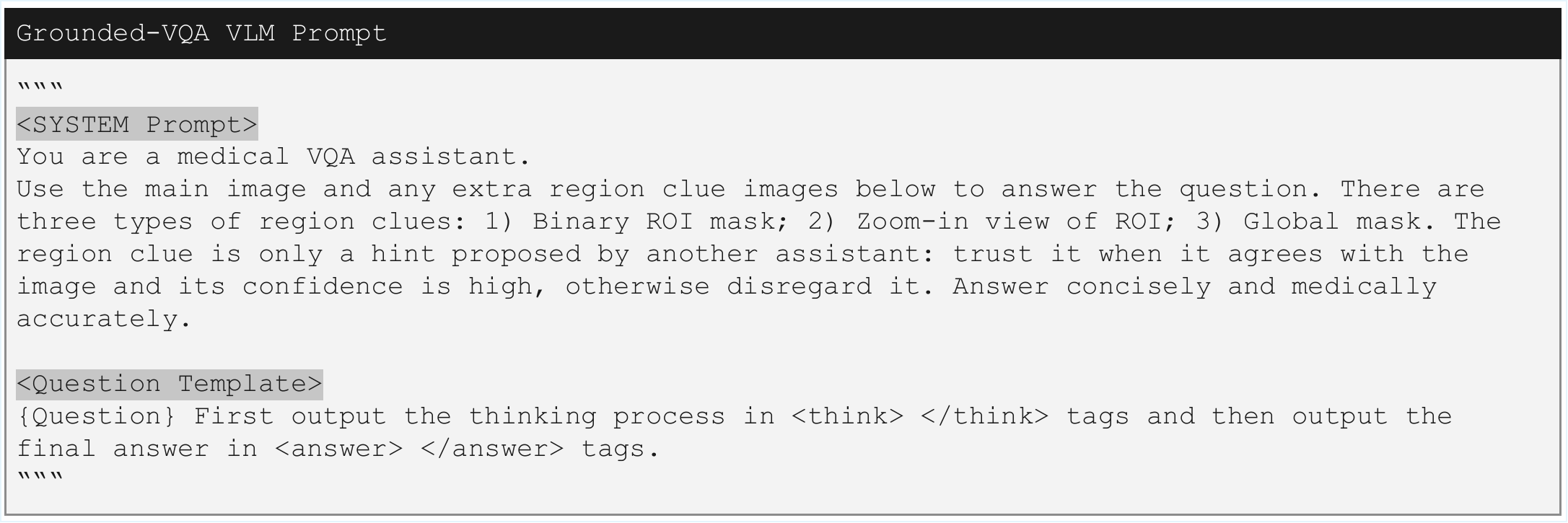}
    \vspace{-3mm}
    \caption{{\bf System Prompt for G-VQA Model.} We present the system prompt for our grounded VQA model, where we provide information about different types of visual clues and guide the model to focus differently when given these visual clues.}
    \vspace{-2mm}
    \label{fig:gvqa_prompt}
\end{figure}
\subsection{EG-VQA Model Prompt}
\label{sec:app_gvqa_prompt}

We present the prompt of the EG-VQA model in \cref{fig:gvqa_prompt}, where we mainly highlight the meaning of each type of visual clue. This can help the model better understand the properties of different visual clues and focus on the corresponding region or local detail when fed with a different clue.

\begin{figure}[!t]
    \centering
    \includegraphics[width=0.8\columnwidth]{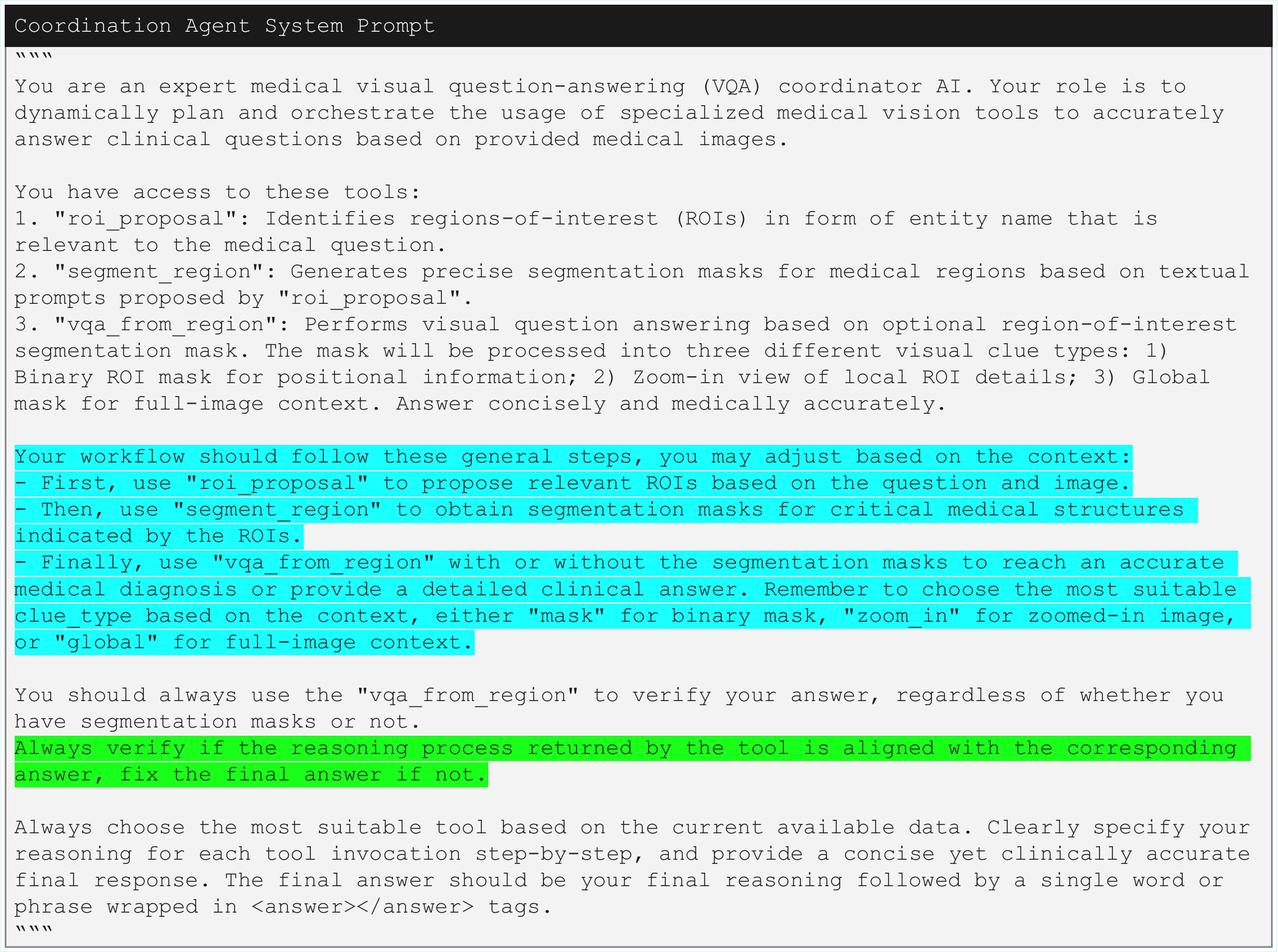}
    \vspace{-3mm}
    \caption{{\bf System Prompt for Coordinator Model.} We present the system prompt for our coordinator model. We introduce the overview of each tool and define the general workflow here. The section highlighted in \hlc[hlcaseblue]{blue} defines the coordination and planning behavior, and the section highlighted in \hlc[hlcasegreen]{green} defines the iterative answer review process during inference.}
    \vspace{-2mm}
    \label{fig:coord_prompt}
\end{figure}

\subsection{Coordinator Prompt}
\label{sec:app_prompt}

We provide the full prompt used for the coordinator in \cref{fig:coord_prompt}. We define the behavior of the coordinator model in the system prompt and instruct it to plan and review. We instruct the coordinator model to review and double-check if the chain of thought and the answer from the tool VLM are consistent, and ask it to correct the answer when necessary. We also require the coordinator to always make at least one tool call to the EG-VQA model, as the EG-VQA model is more convincing on the medical task, and the coordinator model is only for action planning and answer review.

We also provide the tool schema to the coordinator model during inference, which is defined in \cref{sec:app_schema}.

\begin{figure}[!t]
    \centering
    \includegraphics[width=0.8\columnwidth]{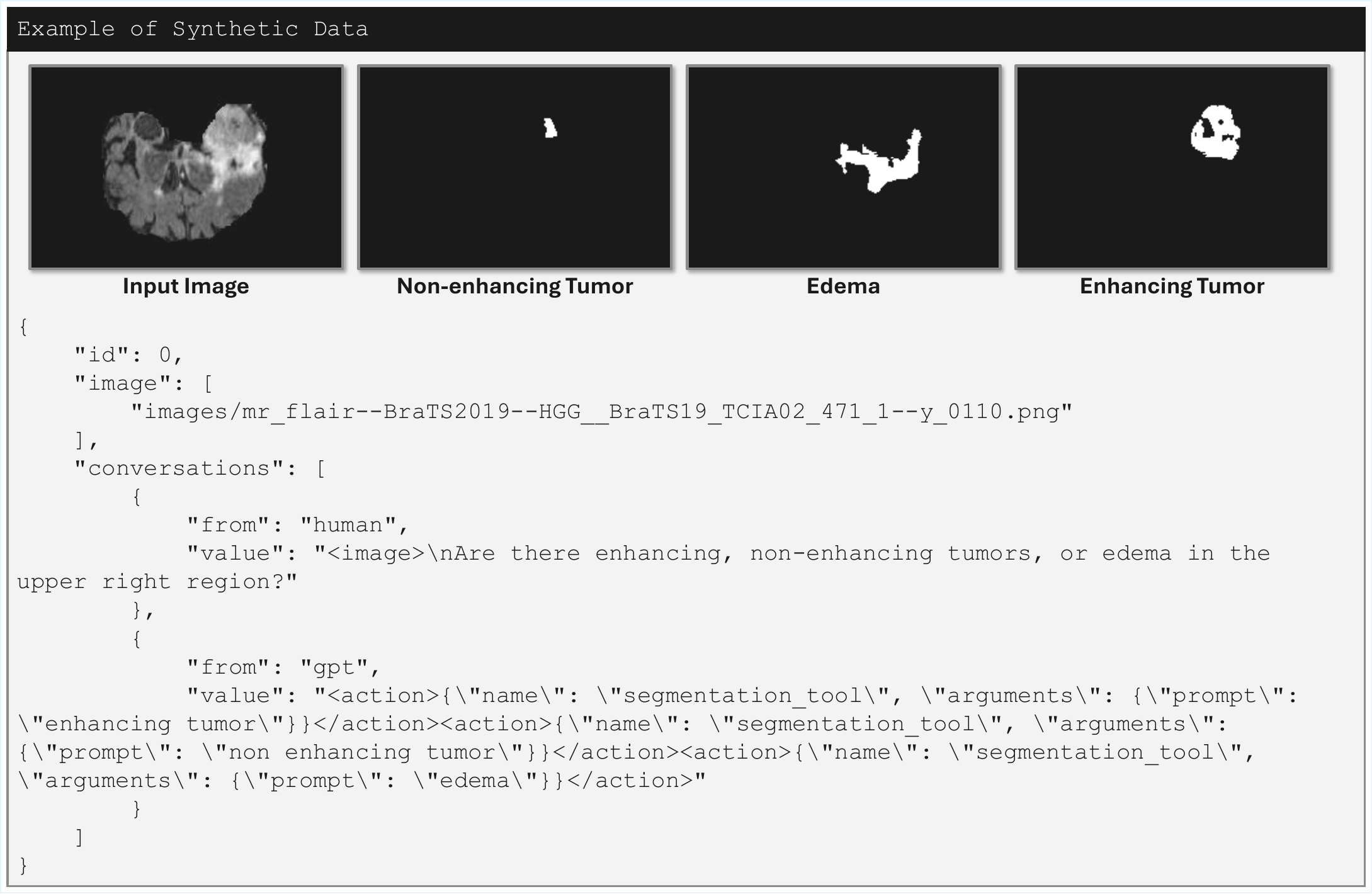}
    \vspace{-1mm}
    \caption{{\bf Example of Synthetic Data for Entity Proposal Task.} An example of synthetic data for the entity proposal task, we wrap the ground truth entity name in the JSON object.}
    \vspace{-2mm}
    \label{fig:data_example1}
\end{figure}
\begin{figure}[!t]
    \centering
    \includegraphics[width=0.8\columnwidth]{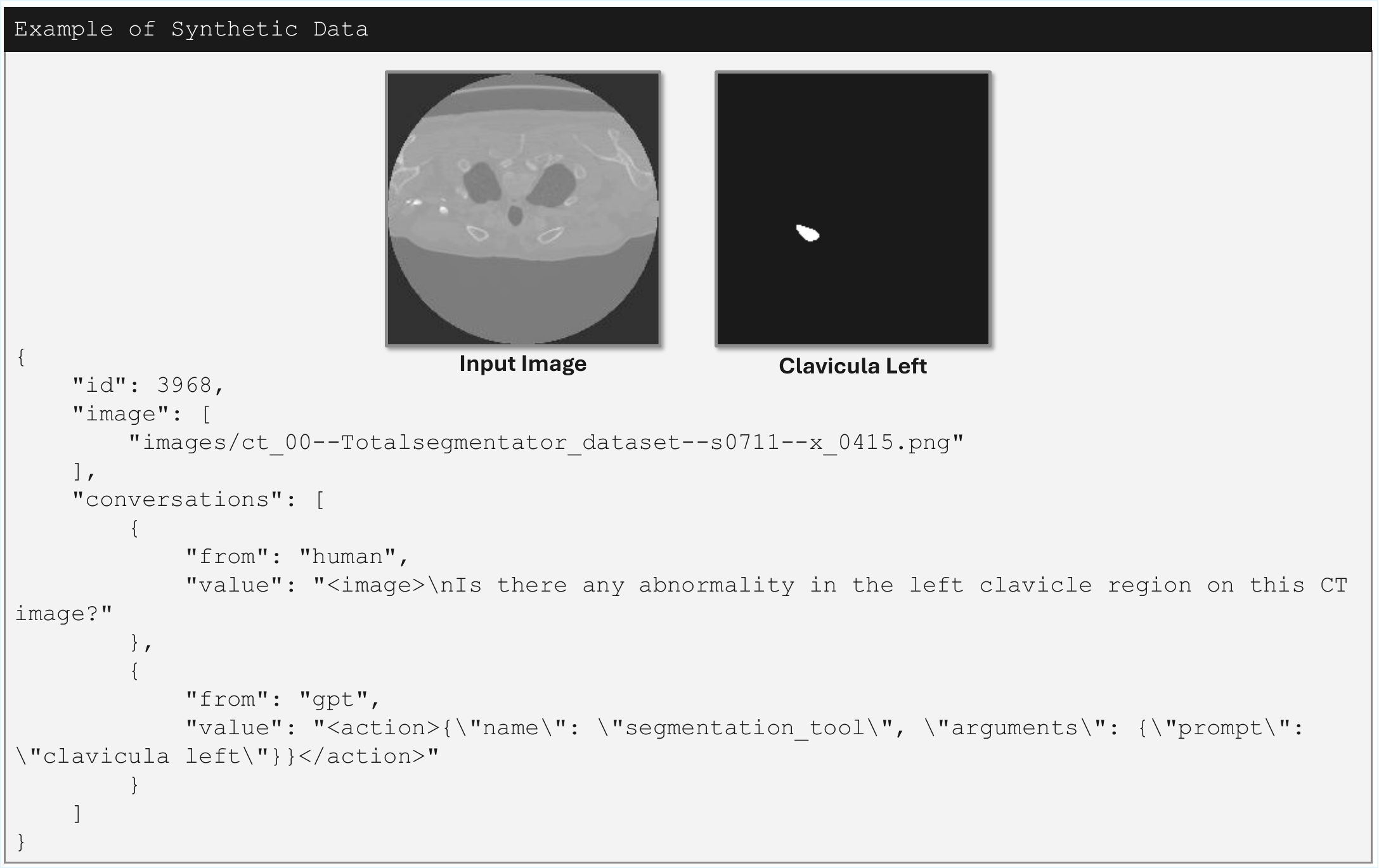}
    \vspace{-3mm}
    \caption{{\bf Example of Synthetic Data for Entity Proposal Task.} An example of synthetic data for the entity proposal task, we wrap the ground truth entity name in the JSON object.}
    \vspace{-2mm}
    \label{fig:data_example2}
\end{figure}
\begin{figure}[!t]
    \centering
    \includegraphics[width=0.8\columnwidth]{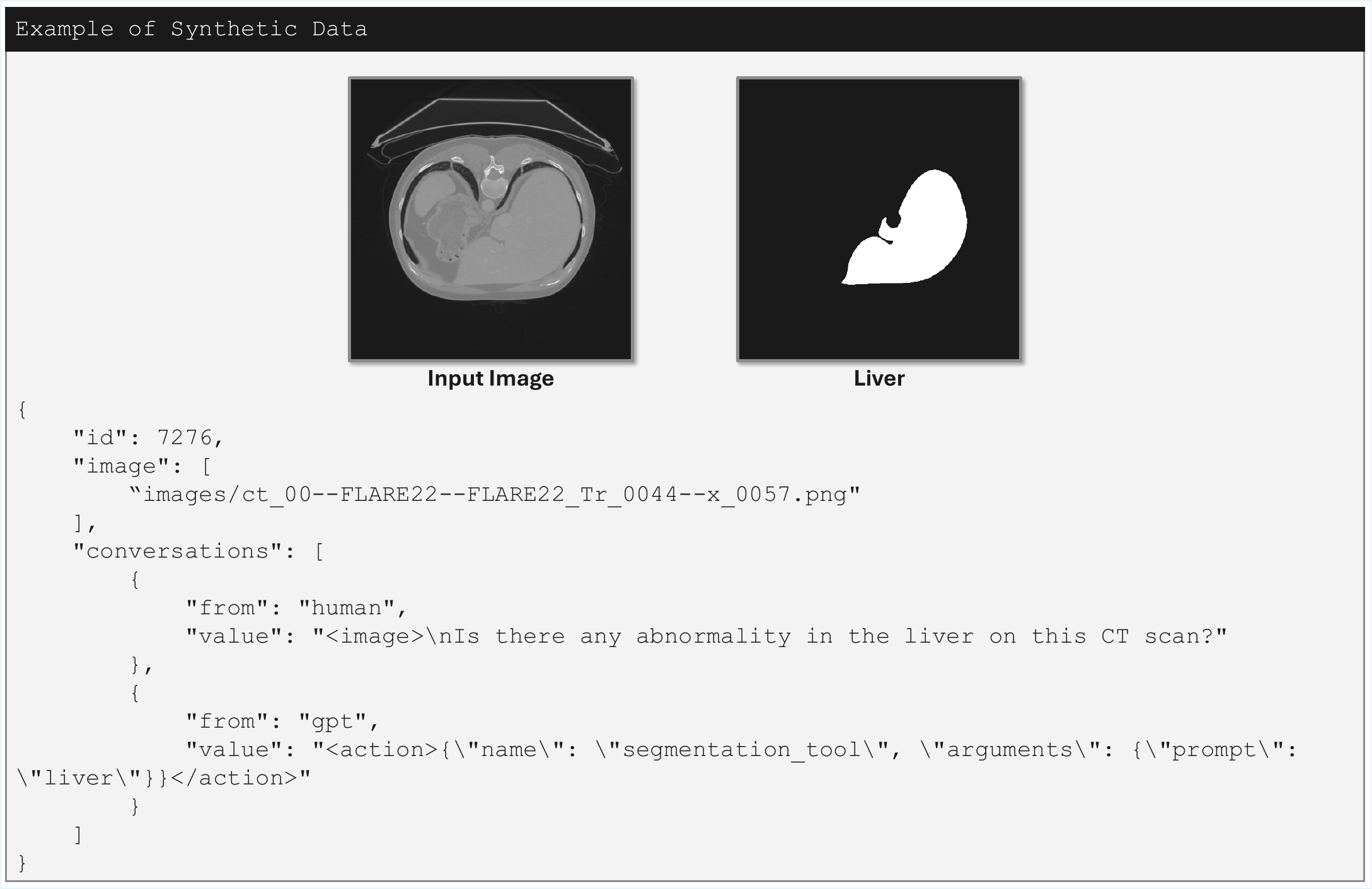}
    \vspace{-3mm}
    \caption{{\bf Example of Synthetic Data for Entity Proposal Task.} An example of synthetic data for the entity proposal task, we wrap the ground truth entity name in the JSON object.}
    \vspace{-2mm}
    \label{fig:data_example3}
\end{figure}
\subsection{Example Synthetic ROI Proposal Data}
\label{sec:app_synth_data_example}

We provide randomly sampled synthetic data examples in \cref{fig:data_example1,fig:data_example2}. The full data will be released upon acceptance.

\subsection{More Case Study}
\label{sec:app_case}

\begin{figure}[!t]
    \centering
    \includegraphics[width=\columnwidth]{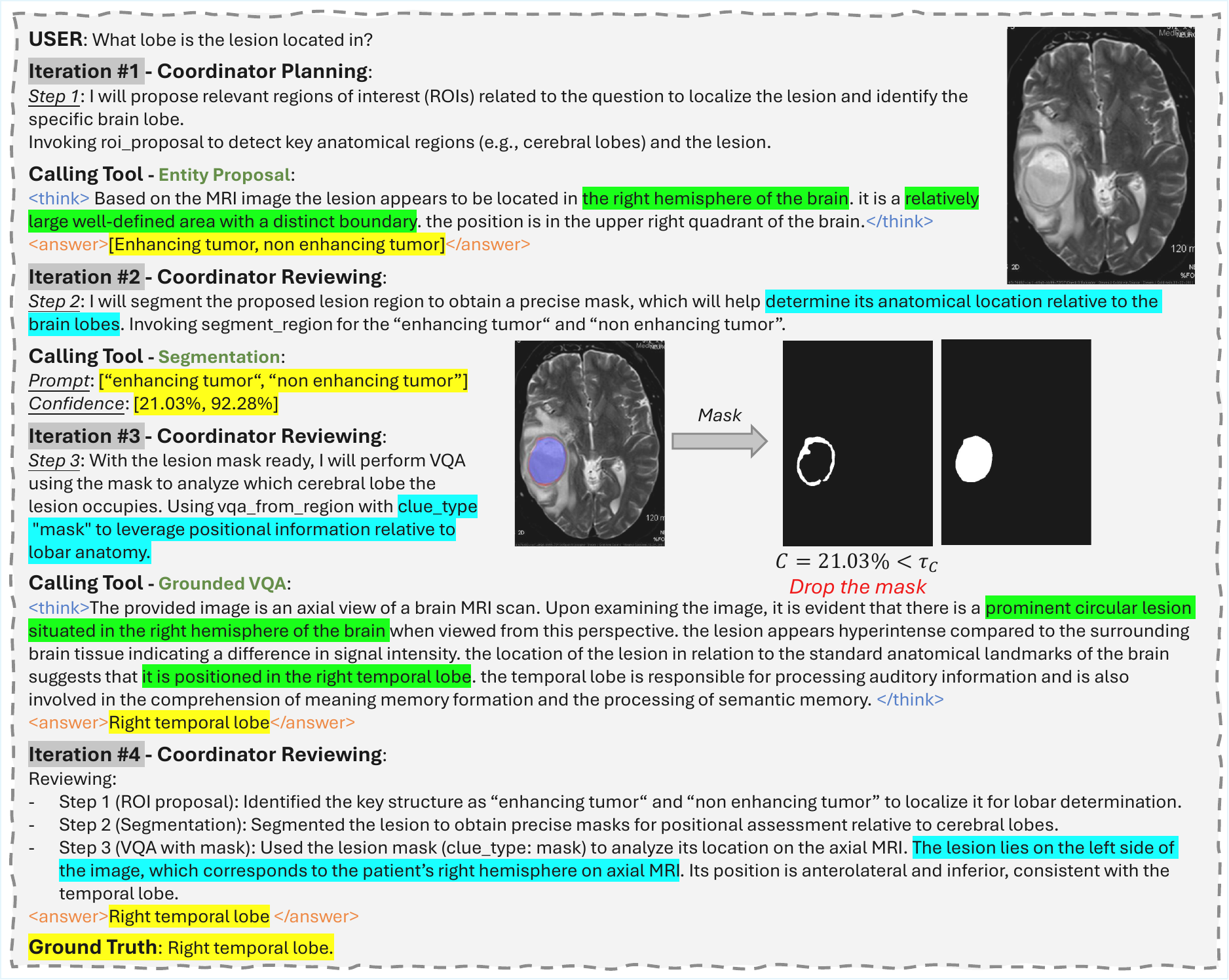}
    \vspace{-3mm}
    \caption{{\bf More Case Study.} We provide a full reasoning process of a user's question about a brain MRI image here. We highlight the key information from the coordinator in blue and key information from the tool in green.}
    \vspace{-2mm}
    \label{fig:case_study1}
\end{figure}

\begin{figure}[!t]
    \centering
    \includegraphics[width=\columnwidth]{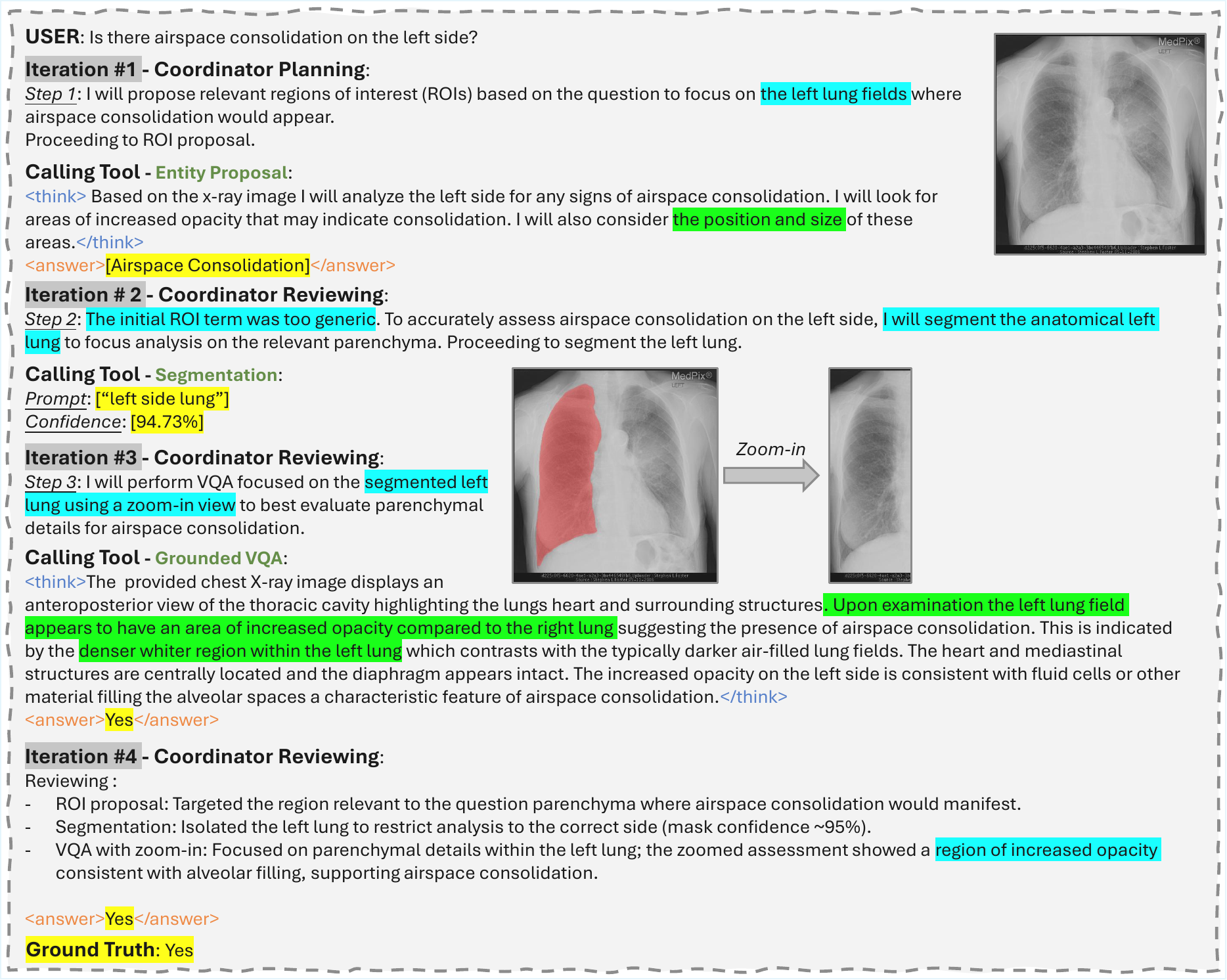}
    \vspace{-3mm}
    \caption{{\bf More Case Study.} We provide a full reasoning process of a user's question about a Chest X-ray image here. We highlight the key information from the coordinator in blue and key information from the tool in green.}
    \vspace{-2mm}
    \label{fig:case_study2}
\end{figure}

\begin{figure}[!t]
    \centering
    \includegraphics[width=\columnwidth]{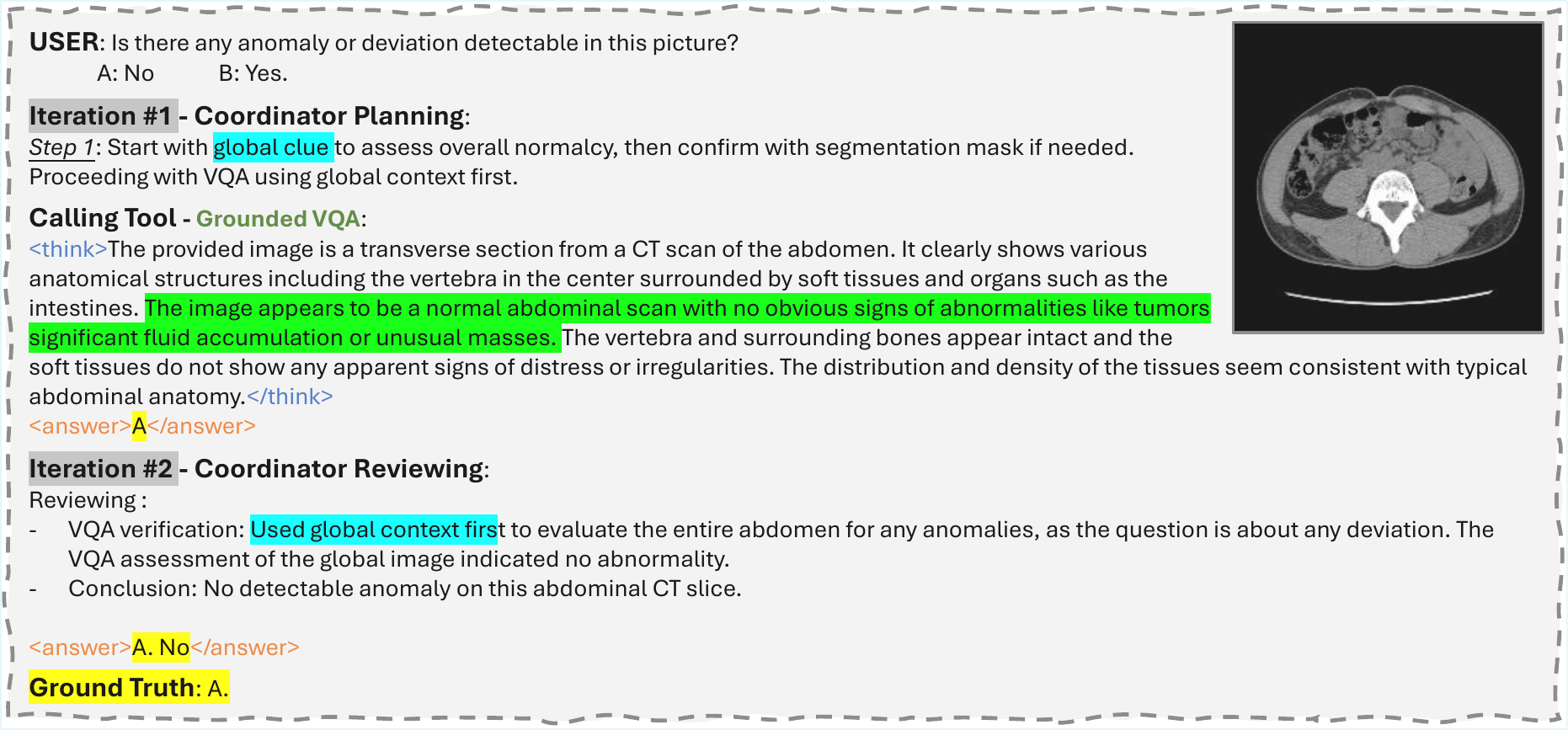}
    \vspace{-3mm}
    \caption{{\bf More Case Study.} We provide a full reasoning process of a user's question about a body CT image here. We highlight the key information from the coordinator in blue and key information from the tool in green. This example skipped the entity proposal and segmentation, as the question is about global information of the image.}
    \vspace{-2mm}
    \label{fig:case_study3}
\end{figure}

\begin{figure}[!t]
    \centering
    \includegraphics[width=\columnwidth]{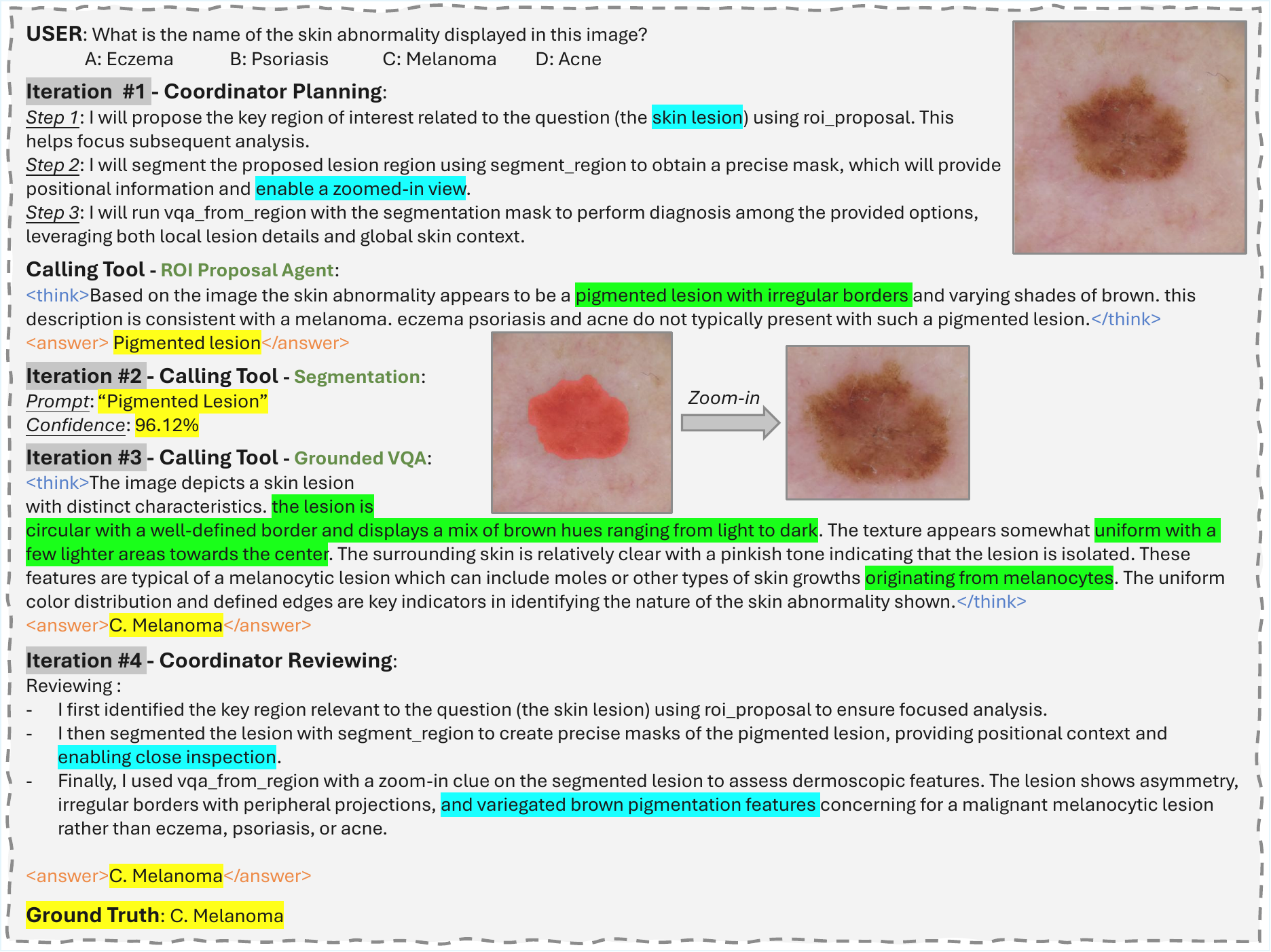}
    \vspace{-3mm}
    \caption{{\bf More Case Study.} We provide a full reasoning process of a user's question about a digital image here. We highlight the key information from the coordinator in blue and key information from the tool in green.}
    \vspace{-2mm}
    \label{fig:case_study4}
\end{figure}

\begin{figure}[!t]
    \centering
    \includegraphics[width=\columnwidth]{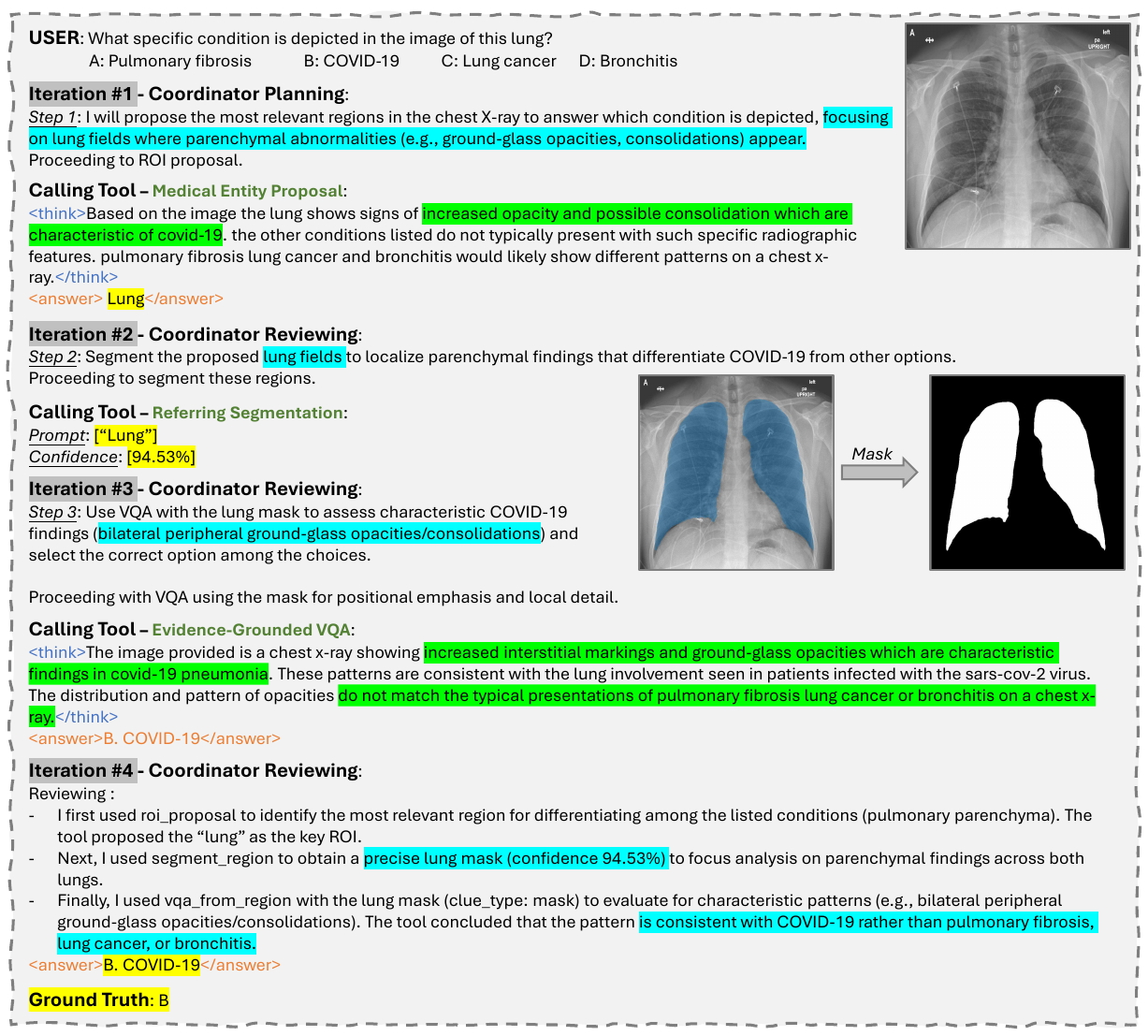}
    \vspace{-3mm}
    \caption{{\bf More Case Study.} We provide a full reasoning process of a user's question about a Chest X-ray here. We highlight the key information from the coordinator in blue and key information from the tool in green.}
    \vspace{-2mm}
    \label{fig:case_study5}
\end{figure}

\begin{figure}[!t]
    \centering
    \includegraphics[width=\columnwidth]{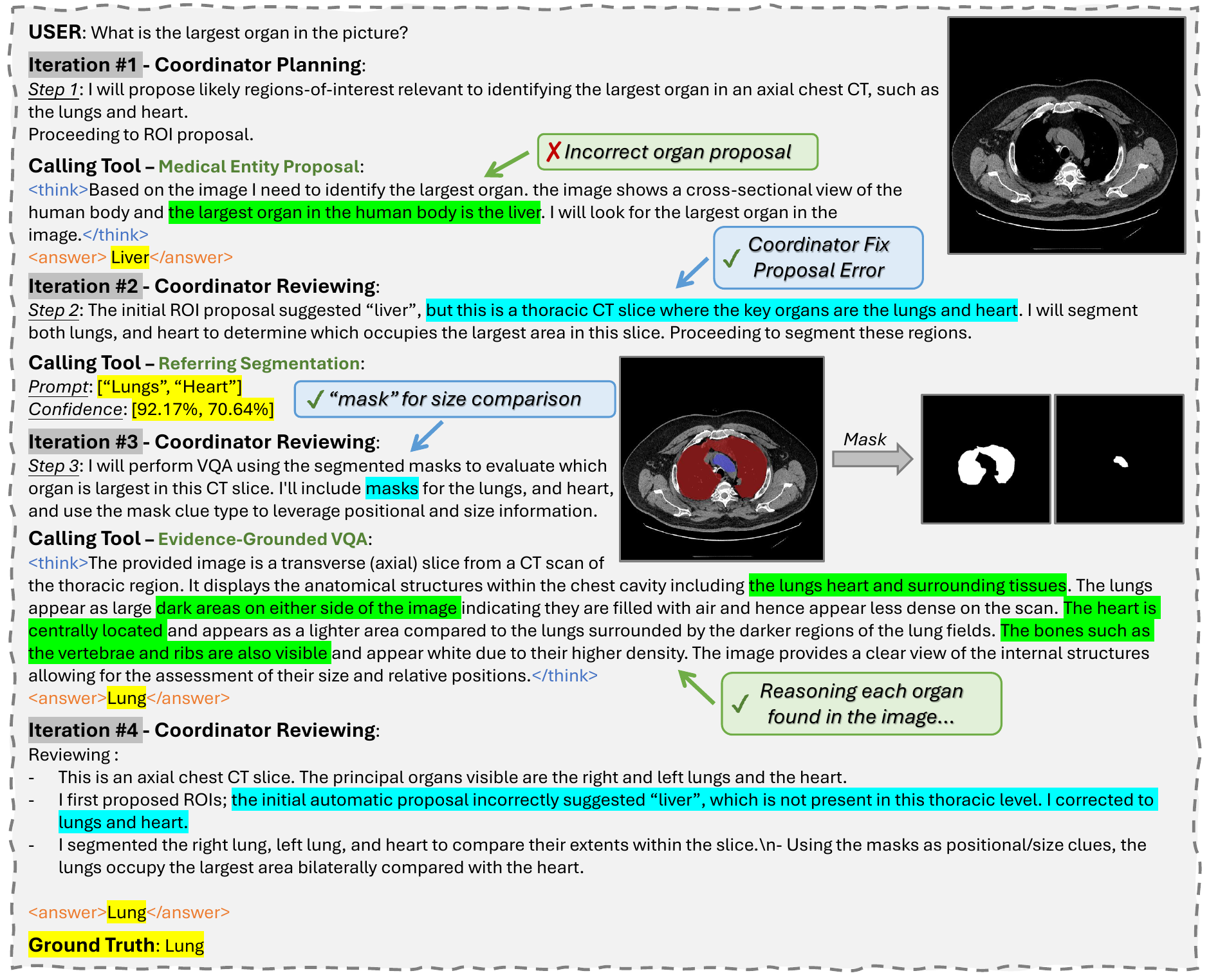}
    \caption{{\bf Case Study.} 
    We present the complete reasoning trace for a CT question.
    Key information from the coordinator is highlighted in \hlc[hlcaseblue]{blue}, model reasoning in \hlc[hlcasegreen]{green}, and each model’s final answer in \hlc[hlcaseyellow]{yellow}. 
    Notably, during review, the coordinator corrects an initial error from the entity proposal.
    }
    \label{fig:case_study6}
\end{figure}

\begin{figure}[!t]
    \centering
    \includegraphics[width=\columnwidth]{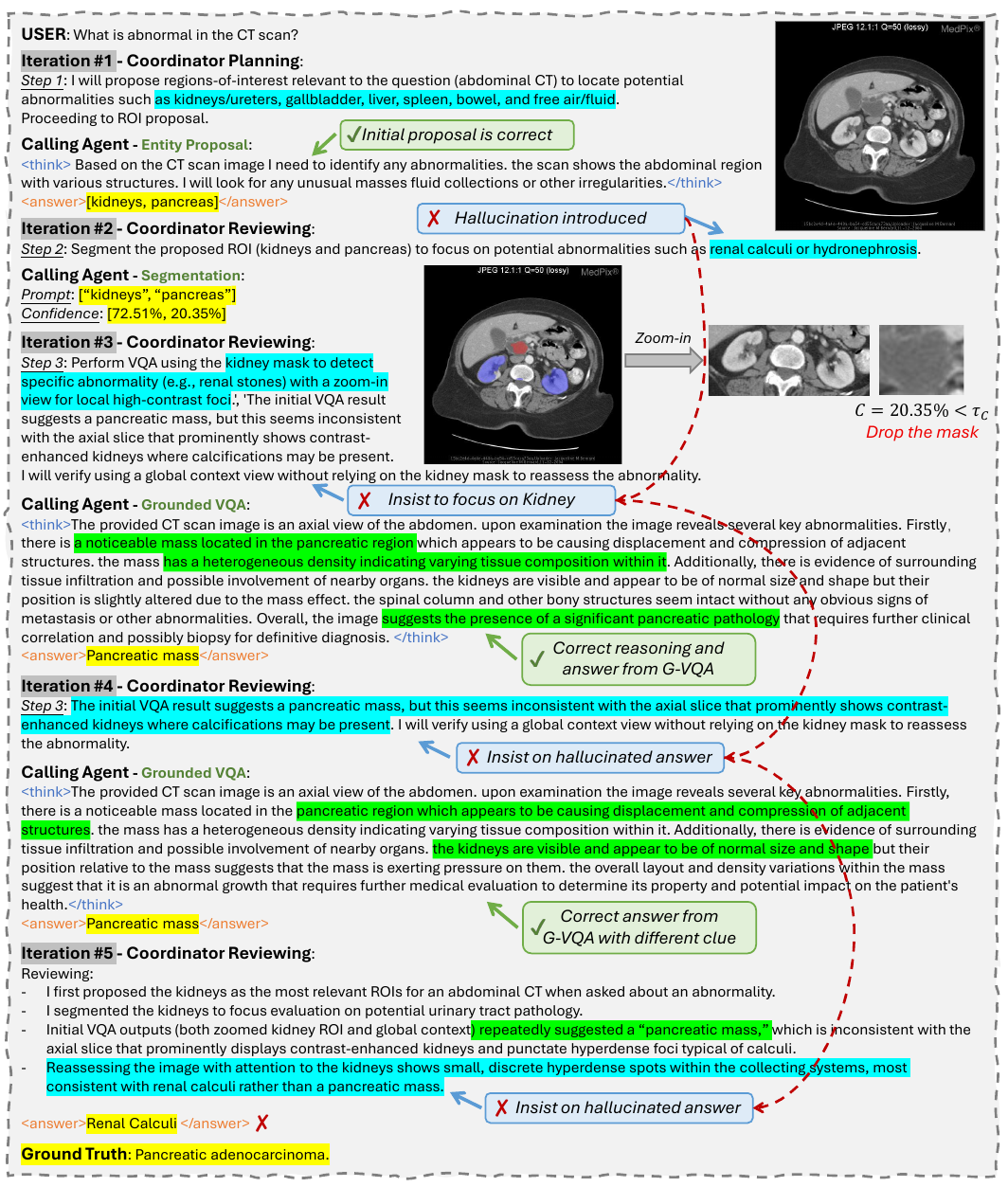}
    \vspace{-3mm}
    \caption{{\bf Failed Case Study.} We provide a full reasoning process of a user's question about a CT image with a failed result here. We highlight the reason for failure during the reasoning chain, where the G-VQA model gives a correct answer, but the coordinator model insists on a hallucinated answer. We use the red dashed arrows to illustrate how the hallucination propagates during the inference.}
    \vspace{-2mm}
    \label{fig:case_study_fail}
\end{figure}

We provide randomly sampled cases during model inference using \name-Coord-B in \cref{fig:case_study1,fig:case_study2,fig:case_study3,fig:case_study4,fig:case_study5,fig:case_study6}. We also include an example of common failure cases in \cref{fig:case_study_fail}, where the reason for failure is due to hallucination introduced by the coordinator. And the coordinator over-edited the answer from the model to insist on the wrong answer.


\subsection{Tool Function Definition}
\label{sec:app_schema}
\begin{figure}[!t]
    \centering
    \includegraphics[width=\columnwidth]{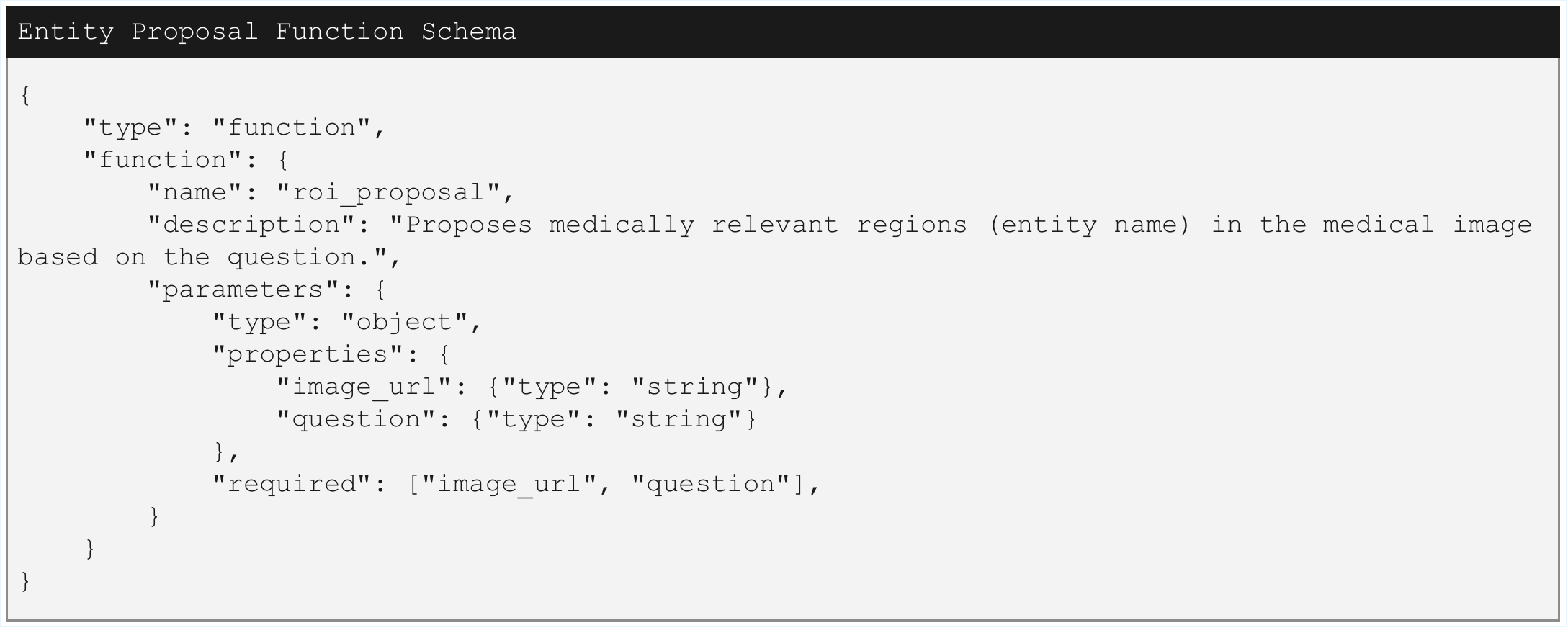}
    \vspace{-3mm}
    \caption{{\bf Function Definition for Entity Proposal Model} We provide the full definition of the function schema of the entity proposal model used during coordination. We explain in detail the definition of each parameter here.}
    \vspace{-2mm}
    \label{fig:schema_entity}
\end{figure}

\begin{figure}[!t]
    \centering
    \includegraphics[width=\columnwidth]{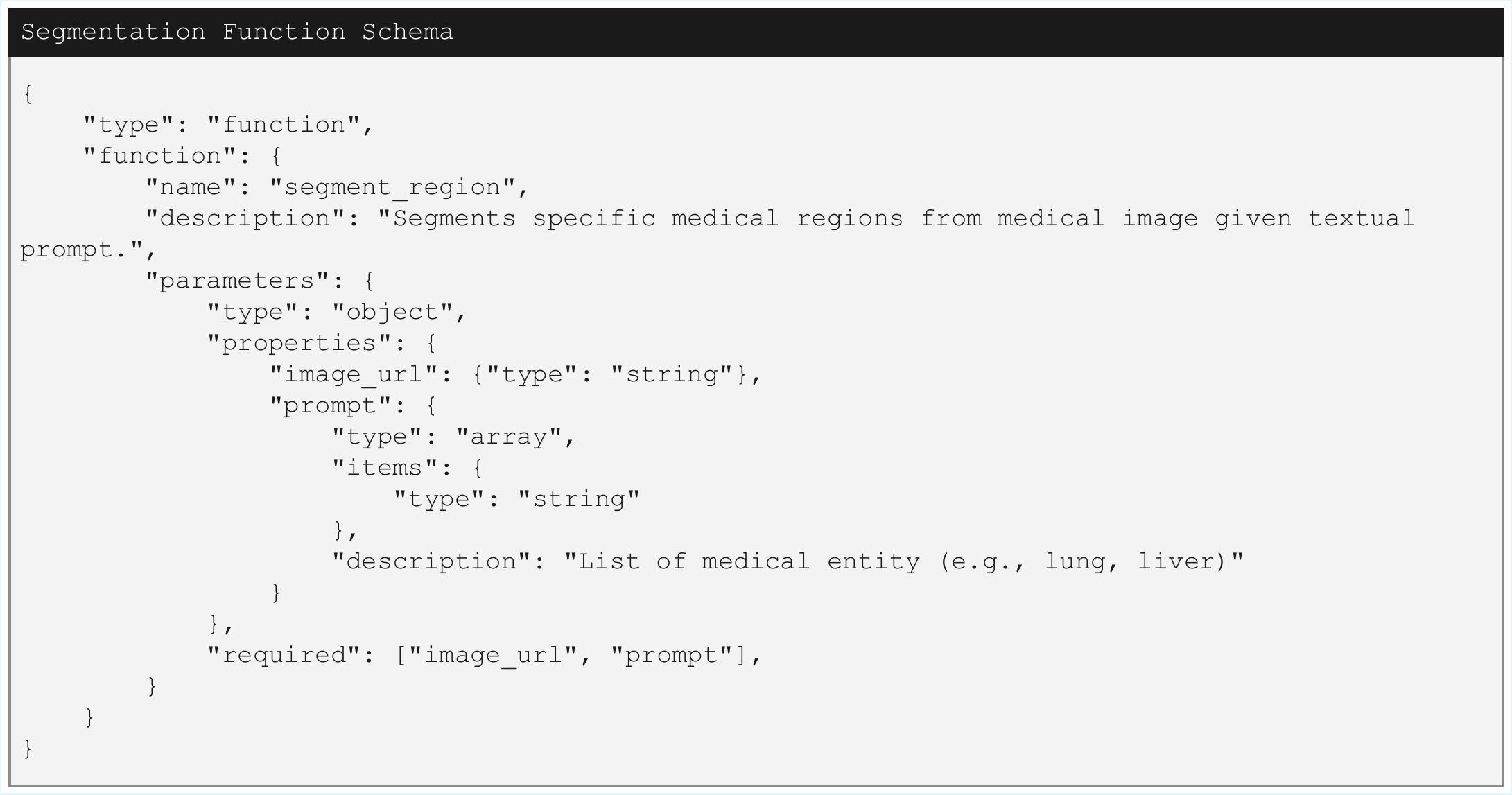}
    \vspace{-3mm}
    \caption{{\bf Function Definition for Segmentation Model} We provide the full definition of the function schema of the referring segmentation model, which accepts two inputs and outputs the segmentation mask along with its confidence.}
    \vspace{-2mm}
    \label{fig:schema_segmentation}
\end{figure}

\begin{figure}[!t]
    \centering
    \includegraphics[width=\columnwidth]{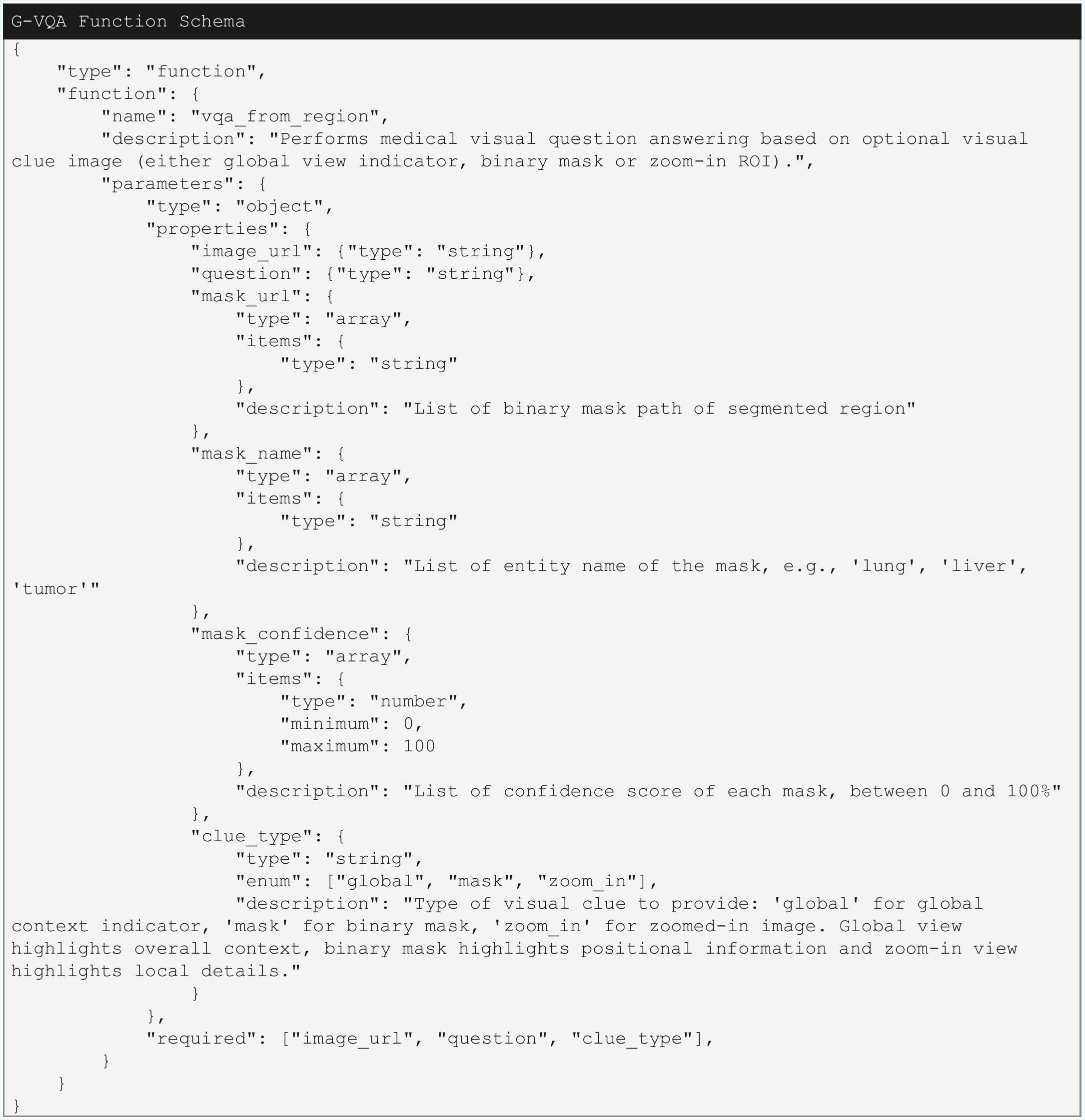}
    \vspace{-3mm}
    \caption{{\bf Function Definition for G-VQA Model} We provide the full definition of the function schema of the G-VQA model. Note that we only set the input image, question, and clue type to be required parameters, as when using a global clue type, there is no need to use other parameters about the mask.}
    \vspace{-2mm}
    \label{fig:schema_gvqa}
\end{figure}

We provide the function definition for the coordinator model in \cref{fig:schema_entity,fig:schema_segmentation,fig:schema_gvqa}.


\end{document}